%% file: main_C.tex
\documentclass[10pt,twocolumn,letterpaper]{article}

\usepackage[pagenumbers]{cvpr} % To force page numbers, e.g. for an arXiv version
\usepackage{multirow}
\usepackage{bbding}
\usepackage{titletoc}
\usepackage{etoc}
\usepackage{tocbibind}
\usepackage{xcolor}
\usepackage{threeparttable}
\usepackage{fontawesome}
\usepackage[accsupp]{axessibility}

\newcommand{\mc}[1]{%
  \begingroup
  \edef\val{#1}%
  \ifdim \val pt > 0pt \textcolor{green!40!black}{#1}%
  \else\ifdim \val pt < 0pt \textcolor{red!60!black}{#1}%
  \else \textcolor{green!40!black}{#1}\fi\fi
  \endgroup
}
\newcommand{\nocontentsline}[3]{}
\let\origcontentsline\addcontentsline
\newcommand\stoptoc{\let\addcontentsline\nocontentsline}
\newcommand\resumetoc{\let\addcontentsline\origcontentsline}

\newcommand\DoToC{%
  \startcontents
  \printcontents{}{1}{\hrule\vskip5pt}
  \vskip5pt\hrule\vskip5pt
}
\input{preamble}
\definecolor{cvprblue}{rgb}{0.21,0.49,0.74}
\usepackage[pagebackref,breaklinks,colorlinks,allcolors=cvprblue]{hyperref}

\def\paperID{730} % *** Enter the Paper ID here
\def\confName{CVPR}
\def\confYear{2026}

\title{TriDF: Evaluating Perception, Detection, and Hallucination\\for Interpretable DeepFake Detection}

\author{
Jian-Yu Jiang-Lin$^1$\thanks{Equal contribution. Author order is alphabetical.} \quad
Kang-Yang Huang$^1$\footnotemark[1] \quad
Ling Zou$^1$\footnotemark[1] \quad
Ling Lo$^2$ \quad
Sheng-Ping Yang$^1$\vspace{0.3em}\\
Yu-Wen Tseng$^1$ \quad
Kun-Hsiang Lin$^1$ \quad
Chia-Ling Chen$^1$ \quad
Yu-Ting Ta$^1$ \quad
Yan-Tsung Wang$^1$\vspace{0.3em}\\
Po-Ching Chen$^1$ \quad
Hongxia Xie$^3$ \quad
Hong-Han Shuai$^2$ \quad
Wen-Huang Cheng$^{1,4}$%\thanks{Corresponding author. \faEnvelopeO~\url{wenhuang@csie.ntu.edu.tw}}
\vspace{0.5em}\\
\centerline{$^1$National Taiwan University \quad $^2$National Yang Ming Chiao Tung University}\vspace{0.3em}\\
\centerline{$^3$Jilin University \quad $^4$VinUniversity}\vspace{0.5em}\\
{\faGithub~\url{https://j1anglin.github.io/TriDF/}}\vspace{-7mm}
}

\input{cvpr26_custom_env}
\newcommand{\benchname}{TriDF}

\begin{document}
\maketitle

\begin{abstract}
Advances in generative modeling have made it increasingly easy to fabricate realistic portrayals of individuals, creating serious risks for security, communication, and public trust. Detecting such person-centric manipulations requires systems not only to distinguish altered content from authentic media but also to provide reliable reasoning. In this paper, we introduce \benchname, a comprehensive benchmark for interpretable DeepFake detection. \benchname\ contains high-quality forgeries from advanced synthesis models, covering 16 DeepFake types across image, video, and audio modalities. The benchmark evaluates three key aspects: Perception, which measures the ability of a model to identify fine-grained manipulation artifacts using human-annotated evidence; Detection, which assesses classification performance across diverse forgery families and generators; and Hallucination, which quantifies the reliability of model-generated explanations. Experiments on state-of-the-art multimodal large language models show that accurate perception is essential for reliable detection, but hallucination can severely disrupt decision-making, revealing the interdependence of these three aspects. \benchname\ provides a unified framework for understanding the interaction between detection accuracy, evidence identification, and explanation reliability, offering a foundation for building trustworthy systems that address real-world synthetic media threats.
\end{abstract}

% \stoptoc
\section{Introduction}
\label{sec:intro}

\begin{figure}[ht]
  \centering
  \includegraphics[width=\linewidth]{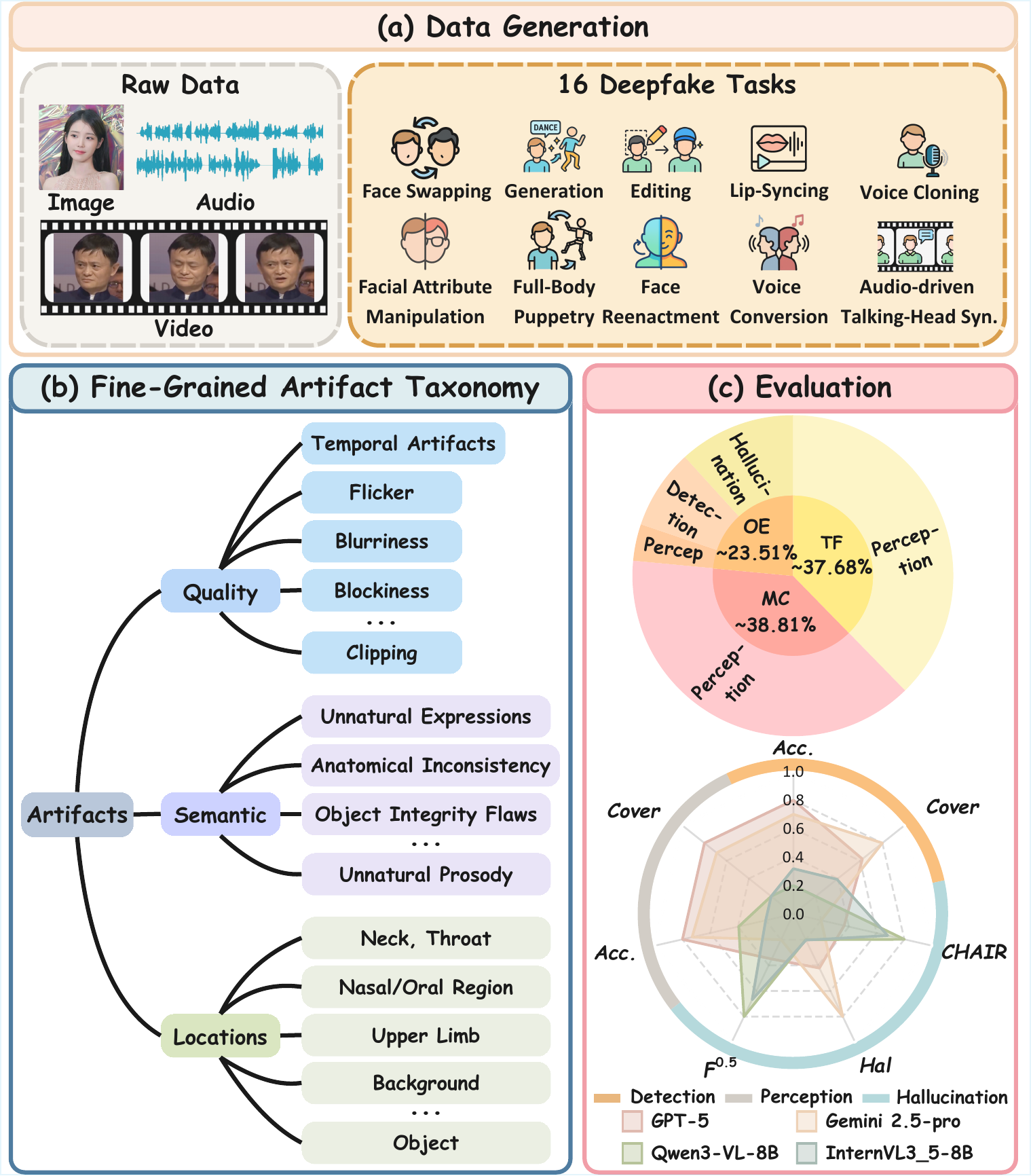} 
  \caption{\textbf{Overview of~\benchname}. We propose~\benchname, a comprehensive benchmark tailored to interpretable DeepFake detection models. (a) We construct $5$K high-quality samples using $16$ DeepFake techniques across three modalities. (b) We design a comprehensive and hierarchical taxonomy of fine-grained artifacts to decompose perception, detection, and hallucination tendency into artifact-wise analyses. (c) The statistics of the proposed~\benchname, and the evaluation results of MLLMs. We normalize the results per metric for clearer comparisons.}
  \label{fig:teaser}
\end{figure}

Fueled by rapid advances in AI-generated content, modern synthesis techniques have intensified the societal risks associated with DeepFakes, a human-centered form of forgery that manipulates or fabricates a person's identity, appearance, or actions. Unlike general synthetic media, DeepFakes specifically target people, creating highly realistic audio, images, and videos that are increasingly difficult to distinguish from genuine human footage. The human-focused nature greatly amplifies their potential for harm, enabling large-scale misinformation campaigns, targeted financial fraud, identity theft, reputational attacks, and severe personal harassment~\cite{luo2021newsclippings,xu2023combating}.

Given the growing threats introduced by recent advances in generative models~\cite{cui2025hallo,peng2024controlnext,liu2025step1x,wu2025omnigen2,wu2025less,wan2025wan,hu2025hunyuancustom}, DeepFake detection has become a critical problem in both research and real-world applications. Beyond simply identifying whether a sample is fake~\cite{smeu2025circumventing, Yang2025D3,yuan2025devil,guo2025face,nam2025m2sformer, huang2024adaptclip}, there is an increasing need for detectors to provide clear and reliable explanations. As Deepfakes directly target human-centered content, stakeholders must understand why a piece of media is considered manipulated rather than relying on an opaque decision. Interpretability is therefore crucial for building trust, enabling human oversight, and supporting accountability in systems that may influence public perception or legal judgments. Moreover, interpretable detection helps reveal which visual, temporal, or acoustic cues modern generators exploit or conceal, offering deeper insight into the evolving landscape of human-centered forgery. As multimodal large language models (MLLMs)~\cite{xu2025fakeshield,huang2025sida,wen2025spot,zhou2025aigi,kang2025legion,guo2025rethinking, Huang2025ThinkFakeRI} become increasingly used for detection~\cite{zou2025survey}, the importance of grounded, human-aligned explanations becomes even more pronounced. 

Despite the increasing importance of explainable deepfake detection, progress is still limited by the shortcomings of current evaluation resources. Previous DeepFake datasets~\cite{rossler2019faceforensics,li2020celeb} have played an important role in advancing raw detection accuracy, yet their annotations are restricted to binary classification. They lack the systematic and fine-grained labels required to evaluate interpretability, and therefore cannot serve as effective benchmarks for modern explainable detection methods. In addition, existing DeepFake benchmarks~\cite{zhang2024common,li2024fakebench,ye2025loki,kang2025legion,zhou2025aigi,huang2025sida,wang2025forensics} suffer from narrow coverage of manipulation types and insufficient generator diversity. As a result, models evaluated using these benchmarks often fail to generalize to the diverse and rapidly evolving landscape of human-centered manipulations. Moreover, a final and critical limitation is the lack of hallucination evaluation in MLLM-based detectors. When these models generate explanations, they may produce incorrect, fabricated, or irrelevant reasoning that does not correspond to any observable artifact in the manipulated sample. Although hallucination metrics have been proposed in other domains~\cite{lin2004rouge}, they are primarily designed for authentic content and do not address the unique challenges posed by DeepFake detection, where explanations must precisely identify manipulation evidence. Without explicit evaluation of hallucination, it is impossible to assess whether an explanation is genuinely grounded in the visual evidence or merely a plausible description that fails to reflect the actual manipulation.

To address the limitations, we introduce Tri-Perspective DeepFake Detection Benchmark, namely \textbf{\benchname}, a comprehensive benchmark designed to evaluate interpretable DeepFake Detection. As shown in~\cref{fig:teaser}, \benchname\ contains high-quality DeepFakes generated by state-of-the-art synthesis models and covers 16 manipulation types across three modalities, including image, video, and audio. The evaluation framework consists of three complementary aspects: \textit{Perception}, \textit{Detection}, and \textit{Hallucination}. \textit{Perception} evaluates whether a model can recognize the manipulation artifacts introduced by different generators. We construct a detailed taxonomy of fine-grained artifact categories, such as quality degradation and semantic inconsistencies, and collect human annotations to establish reliable, human-aligned ground truth. These perceptual labels provide a concrete, structured form of interpretability and allow explanation quality to be assessed in a consistent, evidence-grounded manner. \textit{Detection} measures the ability of a model to distinguish authentic samples from manipulated ones across the full diversity of DeepFake types and generators in \benchname. \textit{Hallucination} evaluates the reliability of model-generated explanations by identifying reasoning that is fabricated or unsupported by the evidence indicated in \textit{Perception}. 

We benchmark a wide range of state-of-the-art MLLMs on \benchname, yielding several important insights. First, accurate perception of manipulation artifacts is a necessary foundation for reliable DeepFake detection. Models that correctly identify fine-grained artifacts tend to perform better in classification, showing that perceiving the right evidence is essential for making correct decisions. However, perception alone is not sufficient. We find that hallucination can severely disrupt detection performance. When a model generates fabricated or unsupported reasoning, its decision-making becomes unstable, and strong perceptual ability no longer translates into accurate detection. The results indicate that detection quality depends jointly on accurate perception and low hallucination. Together, these findings show that perception, detection, and hallucination form an interdependent triad. Neglecting any one of them produces an incomplete picture of the true capability of a detector. The findings underscore the necessity of \benchname, which evaluates all three aspects in an integrated manner and enables a holistic understanding of model reliability in real-world, human-centered DeepFake scenarios.

\begin{figure*}[!ht]
  \centering
  \includegraphics[width=\linewidth]{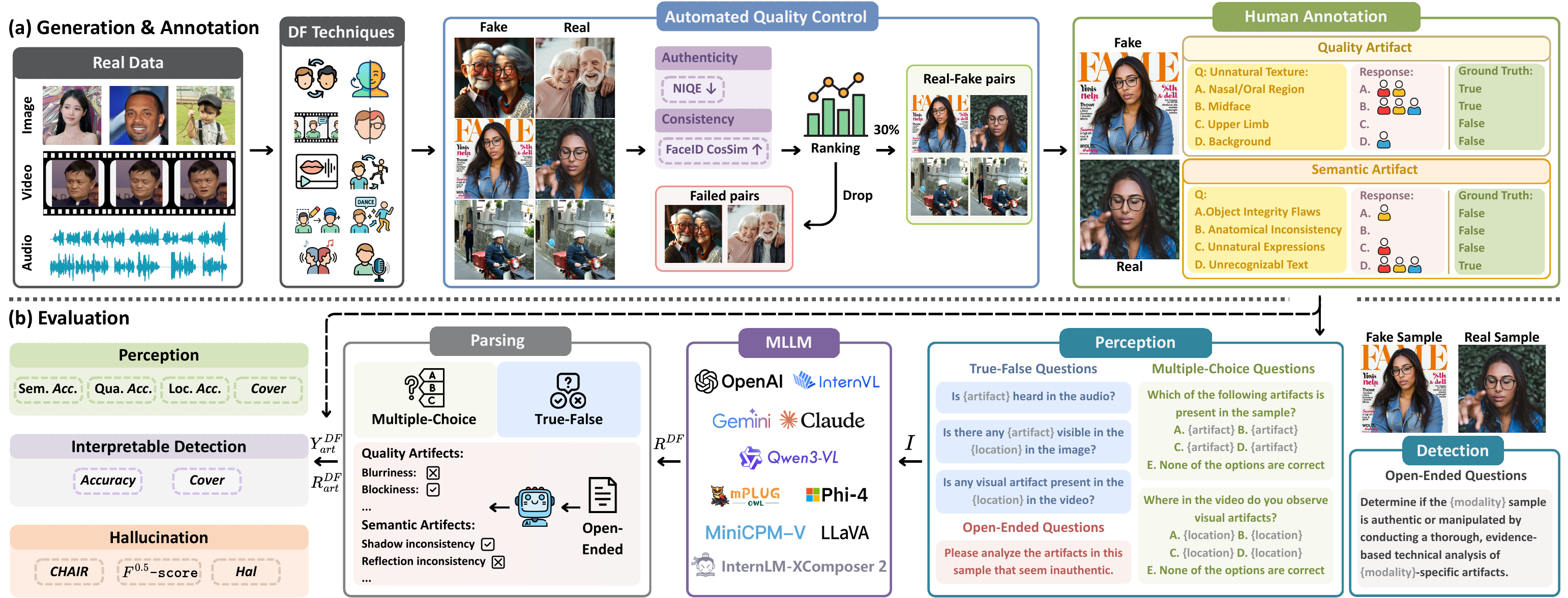}
  \caption{\pp{Pipeline of {\benchname}. (a) Generation \& Annotation: We first collect open-source human-related datasets across three modalities. We generate real-fake data pairs using 16 DeepFake (DF) techniques and perform quality control using authenticity and consistency metrics to obtain high-quality data. We then construct quality and semantic artifact questions and perform human annotation, resulting in reliable ground truth. (b) Evaluation: We design three types of questions, \eg, True-False, Multiple-Choice, and Open-Ended. These questions are combined with high-quality data and fed into MLLMs for evaluation, where the model responses are then assessed using our proposed metrics to evaluate their perception ability, interpretable detection performance, and tendencies towards hallucination.}}
  \label{fig:pipeline}
\end{figure*}

\section{Related Work}
\label{sec:related_work}

\subsection{DeepFake Detection: Trends toward MLLMs}
Conventional DeepFake detection is typically formulated as a supervised binary classification task. Although such models can achieve high accuracy on their training datasets, they often fail to generalize under distribution shifts due to overfitting to dataset-specific cues~\cite{wang2021representative, zhao2021multi, cao2022end, shao2022detecting, yao2023towards}.
Recent image-level approaches incorporate explicit forensic priors and auxiliary objectives that target upsampling traces, frequency artifacts, and cross-view inconsistencies, thereby improving generalization to unseen generators~\cite{Tan2024Upsampling, Liu2024FatFormer, Yang2025D3}. Other methods combine semantic understanding with pixel-level evidence to enhance robustness against high-quality forgeries~\cite{cheng2025co, Nguyen2024LAANet}. For video-based detection, recent advancements incorporate temporal and physiological cues, enforce audio-visual consistency, target challenging facial regions, and utilize training to reduce shortcut reliance~\cite{han2025towards, smeu2025circumventing}. Nevertheless, robustness to unseen manipulations and real-world distortions remains limited.

To enhance generalization and interpretability, MLLM-based detectors combine vision encoders with LLMs for unified detection and reasoning. FakeShield~\cite{xu2025fakeshield}, SIDA~\cite{huang2025sida}, FakeVLM~\cite{wen2025spot}, and KFD~\cite{Yu2025LVLM-DFD} utilize multimodal reasoning and knowledge-guided learning, whereas LEGION~\cite{kang2025legion} and AIGI-Holmes~\cite{zhou2025aigi} emphasize human-like visual and linguistic reasoning, prioritizing conceptual justification over low-level artifacts.

While MLLM-based approaches improve interpretability, their reasoning remains vulnerable to hallucination~\cite{Kalai2025Hallucinate, zou2025survey}. To mitigate this, FFTG~\cite{sun2025fftg} grounds explanations by pairing mask-guided localization from real–fake comparisons with structured prompts and then fine-tuning CLIP and MLLMs via alignment and fusion objectives for more faithful, transferable rationales. Extending to video-level scenarios, AvatarShield~\cite{Xu2025AvatarShield} integrates temporal and semantic reasoning under reinforcement-learning consistency constraints, enhancing interpretability and reducing spurious explanations over time.

\begin{table*}[!ht]
    \centering
    \caption{A comparison of~\benchname~against existing MLLM benchmarks for DeepFake detection. Symbols denote: $\spadesuit$~Accuracy (\eg, F1-score, AUC), $\heartsuit$~Similarity-based (\eg, ROUGE-L, CSS), $\diamondsuit$~LLM-as-a-judge (\eg, GPTScore), and $\clubsuit$~\textit{Cover}.}
    \label{tab:bench}
    \resizebox{\linewidth}{!}{
    \begin{tabular}{l|c|c|c|ccc|c|c|c|c}
        \hline 
         & \textbf{Size of} & \textbf{Number of} & \textbf{DeepFake} & \multicolumn{3}{c|}{\textbf{Data Modality}} & \textbf{Metrics for} & \textbf{Evaluation for} & \textbf{Evaluation for} & \textbf{Real-Fake}\\ 
        \cline{5-7}
        \multirow{-2}{*}{\textbf{Dataset}}  
        & \textbf{Testing Set}
        & \textbf{Generator}
        & \textbf{Types}
        & \textbf{Img} 
        & \textbf{Vid} 
        & \textbf{Aud} 
        & \textbf{Interpretability}
        & \textbf{Perception}
        & \textbf{Hallucination}
        & \textbf{Pair}
        \\ \hline 
        
        % FFHQ
        % FFHQ~\cite{karras2019style} 
        %     % & {[}CVPR'19{]} & 70k & & 1 & \Checkmark & - & - & & \XSolidBrush & \XSolidBrush & \XSolidBrush \\ 

        % FakeSpotter
        % FakeSpotter~\cite{wang2020fakespotter}
        %     % & {[}IJCAI'20{]} & 12k & 7 & 4 & \Checkmark & - & -  & & \XSolidBrush & \XSolidBrush & \XSolidBrush \\ 

        % CNNSpot
        % CNNSpot~\cite{wang2020cnn}
        %     % & {[}CVPR'20{]} & 724k & 11 & 1 & \Checkmark & - & - & & \XSolidBrush & \XSolidBrush & \XSolidBrush \\ 

        % ASVspoof 2019
        % ASVspoof 2019~\cite{wang2020asvspoof} 
        %     & {[}CSL'20{]} & 108k & 2 & - & - & \Checkmark & \XSolidBrush & \Checkmark & \XSolidBrush \\ 

        %% ForgeryNet
        % ForgeryNet~\cite{he2021forgerynet} 
        %     & {[}CVPR'21{]} & 3.1M & 15 & 4 & \Checkmark & \Checkmark & - & & \XSolidBrush & \XSolidBrush & \Checkmark \\         

        % UniversalFake
        % UniversalFake~\cite{ojha2023towards}
        %     % & {[}CVPR'23{]} & 720k & 4 & 1 & \Checkmark & - & - & & \XSolidBrush & \XSolidBrush & \XSolidBrush \\ 

        % Fake2M
        % Fake2M~\cite{lu2023seeing}
        %     % & {[}NeurIPS'23{]} & 2.3M & 11 & 1 & \Checkmark & - & - & & \XSolidBrush & \XSolidBrush & \XSolidBrush \\ 

        % Genimage
        % Genimage~\cite{zhu2023genimage}
        %     % & {[}NeurIPS'23{]} & 2.6M & 8 & 1 & \Checkmark & - & - & & \XSolidBrush & \XSolidBrush & \XSolidBrush \\ 

        % Chameleon
        % Chameleon~\cite{yan2024sanity}
        %     % & {[}ICLR'25{]} & 11k & & 1 & \Checkmark & - & - & & \XSolidBrush & \XSolidBrush & \XSolidBrush \\ 

        % DeepfakeBench
        % DeepfakeBench~\cite{yan2023deepfakebench} 
        %     % & {[}NeurIPS'23{]} & 211k & 24 & 1 & - & \Checkmark & - & \XSolidBrush & \XSolidBrush & \XSolidBrush & \XSolidBrush \\ 
        
        % DD-VQA
        DD-VQA~\cite{zhang2024common}
            % & {[}ECCV'24{]}
            & 15K
            & 4
            & 4
            & \Checkmark & - & - 
            & $\heartsuit$
            & \Checkmark
            & \XSolidBrush 
            & \XSolidBrush
        \\
        % FakeBench
        FakeBench~\cite{li2024fakebench}
            % & {[}IEEE TIF'25{]}
            & 3.6K
            & 10
            & 1
            & \Checkmark & - & - 
            & $\spadesuit$, $\heartsuit$, $\diamondsuit$
            & \XSolidBrush
            & \XSolidBrush 
            & \XSolidBrush
        \\ 

       % FakeClue
        % FakeClue~\cite{wen2025spot}
        %     % & {[}NeurIPS'25{]} & 100k & 17 & 1 & \Checkmark & - & - & & \XSolidBrush & \XSolidBrush & \XSolidBrush \\ 

        % MMFakeBench
        % MMFakeBench~\cite{liu2025mmfakebench}
        %     % & {[}ICLR'25{]} & 11k & 4 & 1 & \Checkmark & - & - & & \XSolidBrush & \XSolidBrush & \XSolidBrush \\ 

        % LEGION 
        SynthScars~\cite{kang2025legion}
            % & {[}ICCV'25{]}
            & 12K
            & 18
            & 4
            & \Checkmark & - & - 
            & $\heartsuit$
            & \XSolidBrush
            & \XSolidBrush 
            & \XSolidBrush
        \\ 

        % LEGION 
        AIGI-Holmes~\cite{zhou2025aigi}
            % & {[}ICCV'25{]}
            & 1K
            & 18
            & 4
            & \Checkmark & - & - 
            & $\heartsuit$, $\diamondsuit$
            & \XSolidBrush
            & \XSolidBrush 
            & \XSolidBrush
        \\ 

        % MMTD-Set
        % MMTD-Set~\cite{xu2025fakeshield}
        %     % & {[}ICLR'25{]} & 153k & 4 & 9 & \Checkmark & - & - & & \XSolidBrush & \XSolidBrush & \XSolidBrush \\ 

        % SID-Set
        SID-Set~\cite{huang2025sida}
            % & {[}CVPR'25{]}
            & 30K
            & 1
            & 1
            & \Checkmark & - & - 
            & $\heartsuit$
            & \XSolidBrush
            & \XSolidBrush 
            & \XSolidBrush
        \\ 

        % HydraFake
        % HydraFake~\cite{tan2025veritas}
            % & {[}ICLR'26{]}
        %    & 50K
        %    & 36
        %    & 8
        %    & \Checkmark & - & - 
        %    & $\spadesuit$
        %    & \XSolidBrush
        %    & \XSolidBrush 
        %    & \XSolidBrush
        % \\ 
        
        % FakeHumanVid
         AvatarShield~\cite{Xu2025AvatarShield}
             % & {[}CVPR'25{]}
             & 15K
            & 9
             & 3
             & - & \Checkmark & - 
             & $\spadesuit$
             & \XSolidBrush
             & \XSolidBrush 
             & \XSolidBrush
        \\ 

        % Forensics-Bench
        Forensics-Bench~\cite{wang2025forensics}
            % & {[}CVPR'25{]}
            & 63K
            & 22
            & 10
            & \Checkmark & \Checkmark & - 
            & $\spadesuit$
            & \XSolidBrush
            & \XSolidBrush 
            & \XSolidBrush
        \\ 

        % LOKI
        LOKI~\cite{ye2025loki}
            % & {[}ICLR'25{]}
            & 18K
            & 18
            & 3 % TTS, VC, image generation,
            & \Checkmark & \Checkmark & \Checkmark 
            & $\spadesuit$, $\diamondsuit$
            & \XSolidBrush
            & \XSolidBrush 
            & \XSolidBrush
        \\ \hline
        
        % Ours
        \benchname~(Ours) 
            % & -
            & \textbf{65K}
            & \textbf{51}
            & \textbf{16}
            & \Checkmark & \Checkmark & \Checkmark 
            & $\spadesuit$, $\clubsuit$
            & \Checkmark
            & \Checkmark 
            & \Checkmark
        \\ \hline

    \end{tabular}
    }
\end{table*}

\subsection{Benchmarks in Deepfake Analysis}
On the benchmarking side, the field has also evolved from early classifier-centric corpora toward benchmarks that emphasize interpretability, multimodality, and reasoning capabilities. Early datasets such as FaceForensics++~\cite{rossler2019faceforensics} and DFDC~\cite{dolhansky2020deepfake} laid the foundation for image-based DeepFake research, while large-scale benchmarks like ForgeryNet~\cite{he2021forgerynet} and LAV-DF~\cite{Cai2023LAVDF} have expanded both modality coverage and supervision granularity. More recently, fully AI-generated suites such as GenImage~\cite{zhu2023genimage} and GenVideo~\cite{chen2024demamba} have further emphasized cross-generator transferability. However, existing datasets and benchmarks have generally lacked explicit consideration of explainability.

To operationalize explainability, several companion datasets have been released alongside detection frameworks. For instance, MMTD-Set~\cite{xu2025fakeshield} and SID-Set~\cite{huang2025sida} integrate pixel-level manipulation masks with natural-language rationales. DD-VQA~\cite{zhang2024common} reformulates facial manipulation forensics as a visual question answering problem equipped with rationale vocabularies, while FakeClue~\cite{wen2025spot} extends analysis across diverse scenarios through artifact-aware textual explanations of synthetic images. Extending to the video modality, FakeHumanVid~\cite{Xu2025AvatarShield} supports temporally aligned reasoning across frames and encompasses multiple video generation conditions. Nonetheless, these datasets remain limited in generative diversity and modality scope, and their rationale annotations, often produced by large language models, may introduce bias or inconsistency.

Recent benchmarks such as FakeBench~\cite{li2024fakebench} explore explainable fake image detection via natural-language annotations and fine-grained forgery taxonomy, evaluating MLLMs on detection, interpretation, and causal reasoning. LOKI~\cite{ye2025loki} further establishes a multimodal benchmark across images, videos, 3D, audio, and text, emphasizing fine-grained anomaly identification and rationalized reasoning to assess interpretability on synthetic content. However, these benchmarks primarily evaluate model outputs instead of confirming whether MLLMs genuinely perceive low-level visual artifacts or reason through high-level semantic inconsistencies. Additionally, their explanatory hallucinations remain unexamined.

\section{\benchname~Benchmark}
\label{sec:benchmark}

\subsection{DeepFake Data Generation}
\label{sec:generation}
To comprehensively assess MLLMs' ability to distinguish DeepFakes from real data, we generate DeepFakes using over $50$ specialized models across more than $30$ public datasets, yielding about $5$K real-synthetic pairs. Given the risks posed by increasingly realistic AI-generated media, we categorize DeepFake generation into two groups: partially manipulated and fully synthetic, covering $16$ tasks in total. Partially manipulated tasks include \textit{face swapping}, \textit{facial attribute manipulation}, \textit{lip-syncing}, \textit{face reenactment}, \textit{full-body puppetry}, \textit{subject-driven editing}, and \textit{voice conversion}. Fully synthetic tasks include \textit{audio-driven talking head synthesis}, \textit{identity-preserving generation}, \textit{text-to-human image/video generation}, \textit{human image-to-video generation}, and \textit{voice cloning}. Details are provided in the supplementary materials.

%\noindent \hky{\textbf{Data Generation}. To comprehensively assess MLLMs' discrimination capabilities, we curate real-synthetic pairs by leveraging specialized models~\cite{chen2024pixart,wu2025omnigen2,liu2025step1x,esser2024scaling,hu2025hunyuancustom,hacohen2024ltx,wan2025wan,nanoBanana,gptImage,veo3} and public datasets~\cite{karras2018progressive,chung2018voxceleb2,rossler2019faceforensics,karras2019style,zen2019libritts,lee2020maskgan,yu2023celebv,chen2024panda,liu2025hoigen} across image, video, and audio modalities. Specifically, we incorporate a wide range of open-source generators with diverse structures, such as generative adversarial networks, Stable Diffusion, and diffusion transformer, along with proprietary ones. To simulate real-world scenarios, real samples from the unseen test set are paired 1:1 with fakes generated by at least three distinct models per technique. This rigorous pairing facilitates precise and fine-grained annotation. Finally, we employ automated quality control using realism and consistency metrics before human annotation. More details are provided in Sec. \textcolor{cvprblue}{B}.} 
\noindent \textbf{Data Generation}. To promote sample diversity, we sourced publicly available real human datasets~\cite{karras2018progressive,chung2018voxceleb2,rossler2019faceforensics,karras2019style,zen2019libritts,lee2020maskgan,yu2023celebv,chen2024panda,liu2025hoigen} spanning image, video, and audio modalities. To accommodate the growing variety of generators, we leverage open-source models such as generative adversarial networks (GAN)-based approaches~\cite{yang2023styleganex}, Stable Diffusion (SD)-based models~\cite{zhao2023diffswap,choi2023diff}, diffusion transformer (DiT)-based models~\cite{chen2024pixart,batifol2025flux}, as well as proprietary ones~\cite{nanoBanana,gptImage,veo3}, all tailored for DeepFake creation to ensure the fidelity and quality in the outputs. For each DeepFake technique, we begin by selecting real samples from test sets or those unused in training to simulate real-world scenarios. We then generate corresponding fake samples using at least three distinct models, forming a multimodal DeepFake dataset with rigorous one-to-one real-fake pairings, enabling precise and fine-grained annotation. Furthermore, we employ specialized metrics to assess realism and consistency, ensuring automatic quality control before initiating the annotation process. More details are provided in the supplementary materials.

\subsection{Fine-Grained Artifact Taxonomy}
\label{sec:annotation}
%\hky{While MLLMs offer potential for grounded reasoning, current benchmarks~\cite{zhang2024common,li2024fakebench,huang2025sida,wen2025spot,xu2025fakeshield,ye2025loki} fail to evaluate the interdependence of perception, reasoning, and hallucination. Furthermore, using MLLMs, such as GPT-4o~\cite{gpt4o}, for automated evaluation introduces self-preference bias~\cite{lu2022fantastically} and inherited limitations~\cite{sun2025fftg}, necessitating more rigorous, human-aligned benchmarking protocols.}
The rapid progression of AI has made DeepFakes increasingly realistic and diverse, creating challenges for both detection and annotation, while exposing the limits of simple real-or-fake labels. Although MLLM-based detectors offer interpretable, anomaly-grounded reasoning, prior work~\cite{zhang2024common,li2024fakebench,huang2025sida,wen2025spot,xu2025fakeshield,ye2025loki} lacks a comprehensive, standardized artifact-annotation framework for jointly evaluating models' perceptual and reasoning abilities and their susceptibility to hallucination. 

Moreover, many benchmarks rely on carefully engineered prompts to leverage powerful MLLMs (\eg, GPT-4o~\cite{gpt4o}) both for generating explanations and judging the outputs of other models, including themselves. Such automated evaluation inherits the limitations and biases of the underlying MLLMs, undermining the reliability of textual explanations~\cite{sun2025fftg} and introducing self-preference bias~\cite{chen2024mllm}.

%\noindent \hky{\textbf{Taxonomy of DeepFake Artifacts}. To enable a more diagnostic evaluation, we introduce a taxonomy that categorizes artifacts into quality and semantic types. \textbf{Quality artifacts} (\eg, blur, flicker) represent localized, low-level irregularities, whereas \textbf{semantic artifacts} (\eg, anatomical flaws, unnatural prosody) necessitate high-level common-sense reasoning. To rigorously assess MLLM localization capabilities, we further ground quality artifacts in specific spatial regions, such as limbs or backgrounds. Detailed taxonomy and annotation protocols are provided in Secs. \textcolor{cvprblue}{C} and \textcolor{cvprblue}{D}.} 
\noindent \textbf{Taxonomy of DeepFake Artifacts}. To address these challenges, we propose a novel taxonomy for assessing DeepFake detectors, aiming to provide a more diagnostic framework. Inspired by~\cite{li2024fakebench,zhang2024common}, our approach categorizes artifacts into two distinct categories based on their nature and the reasoning required to detect them: \textit{quality artifacts} and \textit{semantic artifacts}. \textit{Quality artifacts}, such as blurriness, noise, or flicker, are typically localized issues that can be identified using traditional image processing methods. Conversely, \textit{semantic artifacts}, including anatomical inconsistencies, object integrity flaws, unrecognizable text, or unnatural prosody, require human-like common sense to spot. We further enhance this taxonomy by grounding quality artifacts at specific locations (\eg, the nasal area, limbs, or background) to systematically evaluate the localization abilities of MLLMs. Detailed taxonomy and annotation protocols are provided in the supplementary materials.

\subsection{Benchmark Construction}
\label{sec:construction}
%\noindent \hky{To comprehensively evaluate the abilities of MLLMs, we evaluate MLLM performance across three dimensions: \textbf{Perception}, \textbf{Detection}, and \textbf{Hallucination}, and these assessments leverage True-False Question (\texttt{<TFQ>}), Multiple-Choice Question (\texttt{<MCQ>}), and Open-Ended Question (\texttt{<OEQ>}) formats, combined with distinct parsing strategies tailored to each specific evaluation objective.}
To comprehensively evaluate the abilities of MLLMs, we categorize our assessment into three distinct dimensions: \textit{Perception}, \textit{Detection}, and \textit{Hallucination}. Each dimension employs specific question formats: True-False Questions (\texttt{<TFQ>}), Multiple-Choice Questions (\texttt{<MCQ>}), and Open-Ended Questions (\texttt{<OEQ>}), along with distinct sampling strategies tailored to the evaluation goal. Recognizing that successful DeepFake detection hinges on accurate perception as a foundation for rationalized outcomes, we structure the benchmark to evaluate perceptual acuity, detection proficiency, and the tendency to hallucinate.

%\noindent \hky{\textbf{Perception} evaluates a model's sensitivity to DeepFake artifacts using \texttt{<TFQ>}, \texttt{<MCQ>}, and Type-A \texttt{<OEQ>} formats.}
%\begin{itemize}
    %\item \textbf{Closed-Ended} (\texttt{<TFQ>}, \texttt{<MCQ>}) are categorized into artifact-focused questions, which identify the presence or type of anomalies, and location-focused ones. The latter are subdivided into Type-1 (identifying any artifact in a designated region) and Type-2 (locating a specific artifact). To increase complexity, \texttt{<MCQ>} allow for multiple selections and include a ``none of the above'' option.
    %\item \textbf{Open-Ended} (Type-A \texttt{<OEQ>}) informs models that the sample is a DeepFake, which must then provide a structured, comprehensive analysis of all observable artifacts.
%\end{itemize}
\noindent \textbf{\textit{Perception}} dimension is designed to test the model's sensitivity to DeepFake flaws. It exclusively utilizes manipulated samples across image, video, and audio modalities. This category encompasses \texttt{<TFQ>}, \texttt{<MCQ>}, and Type-A \texttt{<OEQ>}. Within this scope, \texttt{<TFQ>} and \texttt{<MCQ>} are strictly divided into artifact-related questions and location-related questions. Artifact-related questions probe whether a specific anomaly exists or identify which artifacts are present. Location-related questions are further organized into two types: Type-1 asks whether any artifact appears in a designated region or determines its location, while Type-2 queries the presence or location of a specific artifact. To heighten the challenge, each \texttt{<MCQ>} includes a ``none of the above'' option and allows for multiple valid selections. Furthermore, Type-A \texttt{<OEQ>} falls under this perception-focused category, informing the model that the sample is a DeepFake and requiring a comprehensive, structured analysis of all noticeable artifacts under clear headings.

%\noindent \hky{\textbf{Detection} evaluates a model's ability to distinguish authentic from manipulated content using Type-B \texttt{<OEQ>}. Unlike the perception-focused Type-A, Type-B provides no prior ground truth, requiring the model to classify samples independently. To ensure structured and measurable reasoning, the protocol mandates a strict output format: the model must first provide a binary classification, followed by a detailed list of identified artifacts and supporting evidence.}
\noindent \textbf{\textit{Detection}} dimension focuses on the model's capability to distinguish between authentic and manipulated content, necessitating a dataset that contains both real and fake samples. This category relies solely on Type-B \texttt{<OEQ>}. Unlike Type-A, Type-B prompts the model to classify the sample as authentic or manipulated without prior knowledge of the ground truth. The process adheres to explicit guidelines and a strict output format, mandating that the model state its binary decision first, followed by a list of identified artifacts and supporting reasoning.

%\noindent \hky{\textbf{Hallucination} quantifies a model's tendency to fabricate non-existent artifacts. Evaluated via Type-A and Type-B \texttt{<OEQ>} responses for manipulated samples, this metric identifies instances where the model's reasoning includes artifacts that are not present in the ground truth.}
\noindent \textbf{\textit{Hallucination}} dimension evaluates the model's tendency to fabricate non-existent artifacts. This assessment is derived from the responses to both Type-A and Type-B \texttt{<OEQ>} and applies to both real and fake samples to identify instances where the model hallucinates artifacts.

Considering the ``selection bias'' common in MLLMs~\cite{loginova2025addressing,zheng2023large}, we ensure an even distribution of ground truth options. More details are provided in the supplementary material.
%\subsection{Benchmark Statistics}
%\label{sec:stats}

%\noindent \hky{\benchname\ comprises $65$K questions covering $16$ DeepFake techniques across three modalities (image, video, and audio). The dataset includes $5$K real-synthetic pairs encompassing both partially manipulated and fully synthetic content generated by $51$ state-of-the-art models. To evaluate interpretability and hallucination, the benchmark is partitioned into $23$K \texttt{<TFQ>}, $24$K \texttt{<MCQ>}, and $18$K \texttt{<OEQ>}.}

\begin{table*}[!ht]
\centering
\caption{Evaluation of Multimodal DeepFake Perception}
\label{tab:perception}
\renewcommand{\arraystretch}{1.}
\fontsize{6pt}{8pt}\selectfont
{\setlength{\tabcolsep}{0.6mm}
\resizebox{\textwidth}{!}{
\begin{tabular}{l|cccc|cccc|c|c||cccc|cccc|c|c}
\toprule
\multirow{3}{*}{\textbf{MLLM}}
& \multicolumn{10}{c||}{\textbf{\texttt{<TFQ>}}}
& \multicolumn{10}{c}{\textbf{\texttt{<MCQ>}}} \\
\cmidrule(lr){2-11}\cmidrule(lr){12-21}
& \multicolumn{4}{c|}{\textbf{Image}}
& \multicolumn{4}{c|}{\textbf{Video}}
& \multirow{2}{*}{\textbf{Avg.}}
& \multirow{2}{*}{\textbf{Rank}}
& \multicolumn{4}{c|}{\textbf{Image}}
& \multicolumn{4}{c|}{\textbf{Video}}
& \multirow{2}{*}{\textbf{Avg.}}
& \multirow{2}{*}{\textbf{Rank}} \\
\cmidrule(lr){2-5}\cmidrule(lr){6-9}\cmidrule(lr){12-15}\cmidrule(lr){16-19}
& \textit{Semantic} & \textit{Quality} & \textit{Location} & \textit{Avg.}
& \textit{Semantic} & \textit{Quality} & \textit{Location} & \textit{Avg.}
& & 
& \textit{Semantic} & \textit{Quality} & \textit{Location} & \textit{Avg.}
& \textit{Semantic} & \textit{Quality} & \textit{Location} & \textit{Avg.}
& & \\
\midrule

Random Choice             & 50.00\% & 50.00\% & 50.00\% & 50.00\% & 50.00\% & 50.00\% & 50.00\% & 50.00\% & 50.00\% & -- & \mc{0.00} & \mc{0.00} & \mc{0.00} & \mc{0.00} & \mc{0.00} & \mc{0.00} & \mc{0.00} & \mc{0.00} & \mc{0.00} & -- \\
\midrule 
\multicolumn{21}{l}{\textit{Open Source MLLM}} \\
\midrule
InternVL2\_5-8B~\cite{chen2024expanding}        & 57.94\% & 47.87\% & 54.30\% & 53.37\% & 47.55\% & 53.03\% & 53.68\% & 51.42\% & 52.40\% & 12 & \mc{-0.01} & \mc{-0.35} & \mc{0.10} & \mc{-0.09} & \mc{-0.10} & \mc{-0.34} & \mc{-0.05} & \mc{-0.17} & \mc{-0.13} & 16 \\
InternVL2\_5-26B~\cite{chen2024expanding}       & 57.39\% & 48.82\% & 55.63\% & 53.95\% & 47.76\% & 53.72\% & 53.94\% & 51.81\% & 52.88\% & 11 & \mc{0.08} & \mc{-0.12} & \mc{0.22} & \mc{0.06} & \mc{0.08} & \mc{-0.21} & \mc{0.07} & \mc{-0.02} & \mc{0.02} & 9 \\
InternVL2\_5-38B~\cite{chen2024expanding}       & 57.94\% & 48.82\% & 57.07\% & 54.61\% & 47.57\% & 53.83\% & 54.47\% & 51.96\% & 53.28\% & 9 & \mc{0.00} & \mc{-0.21} & \mc{0.23} & \mc{0.01} & \mc{-0.12} & \mc{-0.38} & \mc{-0.07} & \mc{-0.19} & \mc{-0.09} & 14 \\
InternVL3\_5-8B~\cite{wang2025internvl3}        & 56.20\% & 44.91\% & 59.96\% & 53.69\% & 48.76\% & 56.16\% & 57.16\% & 54.03\% & 53.86\% & 7 & \mc{-0.04} & \mc{-0.06} & \mc{0.20} & \mc{0.03} & \underline{\mc{0.18}} & \mc{0.03} & \textbf{\mc{0.20}} & \mc{0.14} & \mc{0.08} & 4 \\
InternVL3\_5-38B~\cite{wang2025internvl3}   & 56.16\% & 51.66\% & 53.24\% & 53.69\% & 45.94\% & 56.07\% & 51.08\% & 51.03\% & 52.36\% & 13 & \underline{\mc{0.13}} & \mc{0.01} & \mc{0.19} & \mc{0.11} & \mc{-0.01} & \mc{-0.16} & \mc{0.08} & \mc{-0.03} & \mc{0.04} & 8 \\
Qwen3-Omni-30B$^\dagger$~\cite{xu2025qwen3}     & 56.87\% & \underline{62.11}\% & 62.52\% & 60.50\% & 50.82\% & \underline{63.13}\% & 60.31\% & \underline{58.09}\% & 59.29\% & 4 & \mc{0.03} & \mc{-0.12} & \underline{\mc{0.28}} & \mc{0.06} & \mc{-0.06} & \mc{-0.14} & \mc{0.07} & \mc{-0.04} & \mc{0.01} & 10 \\
Qwen3-VL-8B-Instruct~\cite{bai2025qwen3}            & 56.87\% & 59.58\% & \underline{64.55}\% & 60.33\% & 48.37\% & 59.26\% & 56.60\% & 54.74\% & 57.54\% & 5 & \mc{0.04} & \mc{-0.16} & \mc{0.18} & \mc{0.02} & \mc{0.07} & \mc{-0.21} & \mc{0.09} & \mc{-0.01} & \mc{0.00} & 11 \\
Qwen3-VL-30B-Instruct~\cite{bai2025qwen3}           & \underline{59.32}\% & 60.49\% & 63.32\% & 61.04\% & 49.04\% & \textbf{67.78}\% & 59.14\% & \textbf{58.65}\% & \underline{59.85}\% & 2 & \mc{0.07} & \mc{0.20} & \textbf{\mc{0.30}} & \mc{0.19} & \mc{0.14} & \underline{\mc{0.23}} & \mc{0.18} & \underline{\mc{0.18}} & \underline{\mc{0.19}} & 2 \\
LLaVA-OV-7B~\cite{li2024llava}            & 39.58\% & 41.25\% & 0.00\% & 26.94\% & 35.57\% & 40.47\% & 0.00\% & 25.35\% & 26.15\% & 18 & \mc{0.05} & \mc{-0.30} & \mc{0.02} & \mc{-0.08} & \mc{-0.02} & \mc{-0.29} & \mc{-0.05} & \mc{-0.12} & \mc{-0.10} & 15 \\
LLaVA-OV-72B~\cite{li2024llava}           & \textbf{61.37}\% & 50.00\% & 56.81\% & 56.06\% & \underline{51.78}\% & 51.84\% & 54.96\% & 52.86\% & 54.46\% & 6 & \mc{0.04} & \mc{0.09} & \mc{0.04} & \mc{0.06} & \mc{0.08} & \mc{0.13} & \mc{0.09} & \mc{0.10} & \mc{0.08} & 5 \\
MiniCPM-V-2.6~\cite{yao2024minicpm}          & 42.30\% & 52.23\% & 45.65\% & 46.73\% & \textbf{52.45}\% & 47.03\% & 46.47\% & 48.65\% & 47.69\% & 16 & \mc{0.04} & \mc{0.06} & \mc{-0.01} & \mc{0.03} & \mc{0.07} & \mc{0.08} & \mc{0.05} & \mc{0.07} & \mc{0.05} & 7 \\
MiMo-VL-7B-SFT~\cite{yue2025mimo}         & 47.39\% & 43.05\% & 37.80\% & 42.75\% & 41.31\% & 49.87\% & 38.17\% & 43.12\% & 42.93\% & 17 & \mc{0.00} & \mc{-0.03} & \mc{0.01} & \mc{-0.01} & \mc{-0.16} & \mc{-0.44} & \mc{-0.20} & \mc{-0.26} & \mc{-0.14} & 17 \\
% * XComposer2.5          & 0.00\% & 0.00\% & 0.00\% & 0.00\% & 0.00\% & 0.00\% & 0.00\% & 0.00\% & 0.00\% & 19 & \mc{0.00} & \mc{0.00} & \mc{0.00} & \mc{0.00} & \mc{0.00} & \mc{0.00} & \mc{0.00} & \mc{0.00} & \mc{0.00} & 14 \\
% * mPLUG-Owl3-7B        & 0.00\% & 0.00\% & 0.00\% & 0.00\% & 0.00\% & 0.00\% & 0.00\% & 0.00\% & 0.00\% & 19 & \mc{0.00} & \mc{0.00} & \mc{0.00} & \mc{0.00} & \mc{0.00} & \mc{0.00} & \mc{0.00} & \mc{0.00} & \mc{0.00} & 14 \\
Idefics2-8B~\cite{laurenccon2024matters}            & 58.06\% & 48.01\% & 55.79\% & 53.95\% & 47.61\% & 53.59\% & 54.61\% & 51.94\% & 52.95\% & 10 & \mc{-0.04} & \mc{-0.05} & \mc{0.12} & \mc{0.01} & \mc{0.08} & \mc{-0.04} & \mc{-0.09} & \mc{-0.02} & \mc{0.00} & 12 \\
Mantis-8B~\cite{jiang2024mantis}              & 56.00\% & 43.02\% & 45.38\% & 48.13\% & 44.99\% & 52.90\% & 54.29\% & 50.73\% & 49.43\% & 15 & \mc{-0.01} & \mc{-0.36} & \mc{0.16} & \mc{-0.07} & \mc{-0.02} & \mc{-0.31} & \mc{0.05} & \mc{-0.09} & \mc{-0.08} & 13 \\
% * Phi-3.5               & 0.00\% & 0.00\% & 0.00\% & 0.00\% & 0.00\% & 0.00\% & 0.00\% & 0.00\% & 0.00\% & 19 & \mc{0.02} & \mc{-0.03} & \mc{0.06} & \mc{0.02} & \mc{0.13} & \mc{-0.04} & \mc{-0.06} & \mc{0.01} & \mc{0.01} & 11 \\
Phi-4$^\ddagger$~\cite{abdin2024phi}                  & 56.00\% & 51.99\% & 55.15\% & 54.38\% & 47.74\% & 56.93\% & 54.79\% & 53.15\% & 53.77\% & 8 & \mc{0.00} & \mc{-0.36} & \mc{0.07} & \mc{-0.10} & \mc{-0.16} & \mc{-0.41} & \mc{-0.11} & \mc{-0.23} & \mc{-0.16} & 18 \\
\midrule 
\multicolumn{21}{l}{\textit{Proprietary Models}} \\
\midrule
GPT-5~\cite{gpt5}                  & 58.10\% & \textbf{68.39}\% & 63.59\% & \textbf{63.36}\% & 48.05\% & 61.43\% & \underline{61.59}\% & 57.02\% & \textbf{60.19}\% & 1 & \mc{-0.01} & \mc{0.07} & \mc{0.18} & \mc{0.08} & \mc{-0.09} & \mc{0.06} & \mc{0.13} & \mc{0.03} & \mc{0.06} & 6 \\
Gemini 2.5-Pro~\cite{comanici2025gemini}         & 57.74\% & 61.50\% & \textbf{65.83}\% & \underline{61.69}\% & 50.78\% & 59.65\% & \textbf{62.30}\% & 57.58\% & 59.63\% & 3 & \mc{0.10} & \mc{0.19} & \mc{0.12} & \underline{\mc{0.14}} & \mc{-0.01} & \mc{0.17} & \underline{\mc{0.19}} & \mc{0.11} & \mc{0.13} & 3 \\
Claude Sonnet 4.5~\cite{Claude}      & 57.74\% & 47.98\% & 54.99\% & 53.57\% & 47.32\% & 52.75\% & 53.07\% & 51.05\% & 52.31\% & 14 & \textbf{\mc{0.16}} & \textbf{\mc{0.29}} & \mc{0.13} & \textbf{\mc{0.19}} & \textbf{\mc{0.23}} & \textbf{\mc{0.28}} & \underline{\mc{0.19}} & \textbf{\mc{0.23}} & \textbf{\mc{0.21}} & 1 \\
\midrule 
FakeShield~\cite{xu2025fakeshield}             & 52.09\% & 56.38\% & 56.75\% & 55.07\% & -- & -- & -- & -- & -- & -- & \mc{0.02} & \underline{\mc{0.24}} & \mc{-0.01} & \mc{0.08} & -- & -- & -- & -- & -- & -- \\
FakeVLM~\cite{wen2025spot}               & 3.16\% & 1.99\% & 0.85\% & 2.00\% & -- & -- & -- & -- & -- & -- & \mc{0.04} & \mc{0.04} & \mc{0.00} & \mc{0.03} & -- & -- & -- & -- & -- & -- \\
%\midrule
%\textbf{Average}       & 45.57\% & 43.05\% & 43.88\% & 44.17\% & 40.64\% & 47.12\% & 44.13\% & 44.38\% & 45.02\% & -- & \mc{0.03} & \mc{-0.04} & \mc{0.11} & \mc{0.03} & \mc{0.02} & \mc{-0.09} & \mc{0.04} & \mc{-0.01} & \mc{0.01} & -- \\
\bottomrule
\end{tabular}}}
\begin{tablenotes}[para,flushleft]
\scriptsize
\parbox{\textwidth}{
\textbf{Notes.} -- indicates unsupported modality. $^\dagger$ implies Qwen3-Omni-30B-A3B-Instruct. $^\ddagger$ denotes Phi-4-multimodal-instruct. \textcolor{green!40!black}{Green} means values $\geq 0$ and \textcolor{red!60!black}{red} means values $< 0$.}
\end{tablenotes}
\end{table*}

\subsection{Evaluation Metric}
\label{sec:metrics}%&\noindent \hky{\textbf{Perception} and \textbf{Detection}. We evaluate \texttt{<TFQ>} using accuracy (Acc.). For \texttt{<MCQ>} with $M$ total options and $K$ ground-truth answers, we award a correct selection $+1/K$ points, while each incorrect selection incurs a $-1/(M-K)$ penalty. Unselected options receive zero points.}
\noindent \textbf{\textit{Perception} and \textit{Detection}}. For \texttt{<TFQ>}, we use accuracy (Acc.) as the evaluation metric. For \texttt{<MCQ>}, each question has $M$ options, with $K$ correct ones. We award $+1/K$ points for each correctly selected option and deduct $1/(M-K)$ points for each incorrectly selected option. Unselected options receive no points, either added or deducted.

%\noindent \hky{As illustrated in~\cref{fig:pipeline}, to handle the unstructured MLLM responses, we utilize an external LLM $\theta$, \eg, Gemini 2.5 Flash-Lite~\cite{geminiFlashLite}, to map free-form text to a binary vector, which avoids the parsing complexities of \texttt{<OEQ>} evaluation by focusing on objective artifact identification rather than subjective scoring, such as GPTScore in~\cite{fu2024gptscore,ye2025loki}. Specifically, for an input $I=\{DF, Que\}$, where $DF$ represents the generated DeepFake sample, and $Que$ denotes the \texttt{<OEQ>}, the model generates an initial response $R^{DF}$, which is then mapped to a predefined set of $n$ artifacts $Art = \{art_{1} \cdots art_{n}\}$ to produce the mapped list $R^{DF}_{art}$:}

%\begin{equation}
%    R^{DF}_{art} = \theta(R^{DF}).
%\end{equation}
%\hky{Based on the mapped artifact list $R^{DF}_{art}$, we define $Y^{DF}_{art}$ as the ground-truth binary vector indicating the presence of artifacts in $DF$. To quantify interpretability, we employ the \textit{Cover} metric~\cite{wang2023amber}, which measures the recall of identified artifacts relative to the ground truth:}

%\begin{equation}
%    \text{\textit{Cover(R)}} = \frac{|R^{DF}_{art} \bigcap Y^{DF}_{art}|}{|Y^{DF}_{art}|}.
%\end{equation}
%\hky{For Type-B \texttt{<OEQ>}, we further report detection Acc. alongside \textit{Cover} to evaluate overall classification performance.}
\noindent Since responses from MLLMs tend to be lengthy and free-form, even with strict instructions or system prompts, we utilize an external large language model (LLM), \eg Gemini 2.5 Flash-Lite~\cite{geminiFlashLite}, to map artifacts. This stable LLM, combined with a simple prompt template (detailed in the supplementary material), produces outputs of either \texttt{yes} or \texttt{no}. Our approach avoids the need for additional parsing in \texttt{<OEQ>} evaluation and differs from methods that rely on powerful closed-source MLLMs as judges, such as GPTScore in~\cite{fu2024gptscore,ye2025loki}. Specifically, \benchname~prompts MLLM with a query, $I=\{DF, Que\}$, where $DF$ represents the generated DeepFake sample, and $Que$ denotes the \texttt{<OEQ>}. As illustrated in~\cref{fig:pipeline}, we obtain the initial response $R^{DF}$ by fitting $I$ into MLLM. We first create an array of predefined artifacts, $Art = \{art_{1} \cdots art_{n}\}$ consisting of $n$ annotated artifacts in~\benchname~to filter unnecessary artifacts in $R^{DF}$. Next, we apply artifact mapping by an external LLM, $\theta$, to $R^{DF}$ to create a mapped artifact list, $R^{DF}_{art} = \{art^{R^{DF}}_{1} \cdots art^{R^{DF}}_{n}\}$, defined as:
\begin{equation}
    R^{DF}_{art} = \theta(R^{DF}).
\end{equation}
After obtaining the mapped artifact list $R^{DF}_{art}$, we further construct $Y^{DF}_{art}$, which is a list where values indicate positive or negative presence in the input $DF$. This allows us to quantify the interpretability of DeepFake detection by calculating \textit{Cover}~\cite{wang2023amber} using $R^{DF}_{art}$ and $Y^{DF}_{art}$ to measure the coverage of artifacts in the response, defined as:
\begin{equation}
    \text{\textit{Cover(R)}} = \frac{|R^{DF}_{art} \bigcap Y^{DF}_{art}|}{|Y^{DF}_{art}|}.
\end{equation}
For Type-B \texttt{<OEQ>}, we further report accuracy (Acc.) to evaluate the detection performance, in addition to \textit{Cover}.

%\noindent \hky{\textbf{Hallucination}. To evaluate hallucination tendencies, we adapt \textit{CHAIR}~\cite{rohrbach2018object}, \textit{Hal}~\cite{wang2023amber}, and \textit{$F$-score}~\cite{li2023evaluating}. \textit{CHAIR} quantifies the frequency of hallucinated artifacts in $R^{DF}_{art}$:}

%\begin{equation}
    %\text{\textit{CHAIR(R)}} = 1 - \frac{|R^{DF}_{art} \bigcap Y^{DF}_{art}|}{|R^{DF}_{art}|}.
%\end{equation}
%\hky{\textit{Hal} denotes whether $R^{DF}_{art}$ contains at least one hallucination, where $\text{\textit{Hal(R)}} =1$ if \textit{CHAIR} $> 0$, and $0$ otherwise.}
\noindent \textbf{\textit{Hallucination}}. Drawing from prior works~\cite{guan2024hallusionbench,liu2025phd}, we resort to \textit{CHAIR}~\cite{rohrbach2018object}, \textit{Hal}~\cite{wang2023amber}, and \textit{$F$-score}~\cite{li2023evaluating} to assess the hallucination tendencies of MLLMs. \textit{CHAIR} is a widely used metric measuring the frequency of hallucinatory artifacts appearing in responses. It can be calculated as:
\begin{equation}
    \text{\textit{CHAIR(R)}} = 1 - \frac{|R^{DF}_{art} \bigcap Y_{art}|}{|R^{DF}_{art}|}.
\end{equation}
\textit{Hal} represents the percentage of responses containing hallucinations, defined as
\begin{equation}
    \text{\textit{Hal(R)}} = 
    \begin{cases}
        1 &\quad \mathrm{if}~\text{\textit{CHAIR(R)}} \neq 0 \;\\
        0 &\quad \mathrm{otherwise}. \\
    \end{cases}
\end{equation}

%\noindent \hky{To address the impact of hallucinations on precision, we adopt \textit{$F^{0.5}$-score}~\cite{kaul2024throne} that treats precision as twice as important as recall, which can be formulated as:}
%\begin{equation}
%    \mathbf{F}^{0.5}(R) = \frac{1.25 \cdot (1-\text{\textit{CHAIR(R)}}) \cdot \text{\textit{Cover(R)}}}{(0.25 \cdot (1-\text{\textit{CHAIR(R)}}))+ \text{\textit{Cover(R)}}},
%\end{equation}
%\hky{We assign a penalty of \textit{CHAIR} $=1$ if the model fails to identify any artifact or incorrectly classifies a DeepFake as real. All metrics are computed on a per-sample basis.}
To account for false positives, which are often driven by hallucinations and can severely impact precision, we follow THRONE~\cite{kaul2024throne} and weight precision twice as important as recall, resulting in the \textit{$F^{\beta}$-score}. It can be formulated as:
\begin{equation}
    F^{\beta}(R) = \frac{(1+\beta^2) \cdot (1-\text{\textit{CHAIR(R)}}) \cdot \text{\textit{Cover(R)}}}{(\beta^2 \cdot (1-\text{\textit{CHAIR(R)}}))+ \text{\textit{Cover(R)}}},
\end{equation}
where $\beta$ is $0.5$.

In cases where the list of mapped artifacts has a length of $0$, we assign a value of $1$ to \textit{CHAIR} as a penalty. This indicates that the MLLM has failed to properly address the \texttt{<OEQ>}. Similarly, if the model mistakenly classifies a fake sample as real, we also set \textit{CHAIR} to $1$. All the metrics are computed on a per-sample basis. Additional details are provided in the supplementary material.

\begin{table*}[!t]
\centering
\caption{Evaluation of Interpretable DeepFake Detection, Perception and Hallucination Robustness}
\label{tab:open-ended}
\renewcommand{\arraystretch}{1.}
\fontsize{6pt}{8pt}\selectfont
{\setlength{\tabcolsep}{0.6mm}
\resizebox{\textwidth}{!}{
\begin{tabular}{l|cccc|cccc||ccccc|ccccc}
\toprule
\multirow{3}{*}{\textbf{MLLM}} 
& \multicolumn{8}{c||}{\textbf{\texttt{Type-A <OEQ>}}} 
& \multicolumn{10}{c}{\textbf{\texttt{Type-B <OEQ>}}} \\
\cmidrule(lr){2-9}\cmidrule(lr){10-19}
& \multicolumn{4}{c|}{\textbf{Image}} 
& \multicolumn{4}{c||}{\textbf{Video}} 
& \multicolumn{5}{c|}{\textbf{Image}} 
& \multicolumn{5}{c}{\textbf{Video}} \\
\cmidrule(lr){2-5}\cmidrule(lr){6-9}\cmidrule(lr){10-14}\cmidrule(lr){15-19}
& \textit{Cover} $\uparrow$ & \textit{CHAIR} $\downarrow$ & \textit{Hal} $\downarrow$ & $\mathbf{F}^{0.5}$ $\uparrow$
& \textit{Cover} $\uparrow$ & \textit{CHAIR} $\downarrow$ & \textit{Hal} $\downarrow$ & $\mathbf{F}^{0.5}$ $\uparrow$
& \textit{Acc.} & \textit{Cover} $\uparrow$ & \textit{CHAIR} $\downarrow$ & \textit{Hal} $\downarrow$ & $\mathbf{F}^{0.5}$ $\uparrow$
& \textit{Acc.} & \textit{Cover} $\uparrow$ & \textit{CHAIR} $\downarrow$ & \textit{Hal} $\downarrow$ & $\mathbf{F}^{0.5}$ $\uparrow$ \\
%\multicolumn{19}{l}{\textit{Open Source MLLM}}\\
\midrule
\multicolumn{19}{l}{\textit{Open Source MLLM}} \\
\midrule
InternVL2\_5-8B~\cite{chen2024expanding} & 0.4162 & \textbf{0.5260} & 0.9090 & 0.4332 & 0.2452 & \underline{0.5906} & 0.9489 & 0.3345 & 0.5166 & 0.1670 & 0.8479 & 0.9973 & 0.1531 & \underline{0.5996} & \textbf{0.2276} & \textbf{0.7275} & 0.9950 & \textbf{0.2541} \\
InternVL2\_5-26B~\cite{chen2024expanding} & 0.5130 & 0.5869 & 0.9845 & 0.4152 & 0.2325 & 0.7216 & 0.9913 & 0.2547 & 0.4800 & 0.0921 & 0.9304 & 0.9993 & 0.0745 & 0.3405 & 0.0029 & 0.9972 & 1.0000 & 0.0104 \\
InternVL2\_5-38B~\cite{chen2024expanding} & 0.4781 & 0.5570 & 0.9602 & \underline{0.4342} & 0.2581 & 0.6772 & 0.9571 & 0.2879 & 0.5747 & 0.2306 & 0.8066 & 0.9993 & 0.1971 & 0.5790 & 0.1778 & \underline{0.7423} & \textbf{0.9151} & \underline{0.2152} \\
InternVL3\_5-8B~\cite{wang2025internvl3} & 0.4255 & 0.5750 & 0.9130 & 0.4031 & 0.2934 & 0.6645 & 0.9822 & 0.3077 & 0.4176 & 0.0270 & 0.9745 & 1.0000 & 0.0296 & 0.4722 & 0.0803 & 0.9136 & 0.9991 & 0.0871 \\
InternVL3\_5-38B~\cite{wang2025internvl3} & 0.3462 & 0.6800 & 0.9945 & 0.3144 & 0.2323 & 0.6574 & 0.9657 & 0.2946 & 0.4980 & 0.0482 & 0.9538 & 1.0000 & 0.0455 & 0.4118 & 0.0308 & 0.9725 & 0.9995 & 0.0314 \\
Qwen3-Omni-30B-A3B-Instruct~\cite{xu2025qwen3} & 0.4991 & 0.5697 & 0.9582 & 0.4232 & 0.2550 & 0.6426 & 0.9370 & 0.2975 & \underline{0.6942} & \underline{0.4143} & 0.6701 & 1.0000 & \underline{0.3381} & 0.5146 & 0.1717 & 0.8487 & 0.9977 & 0.1504 \\
Qwen3-VL-8B-Instruct~\cite{bai2025qwen3} & 0.3499 & 0.6597 & 0.9845 & 0.3378 & 0.1702 & 0.7707 & 0.9881 & 0.2083 & 0.6207 & 0.2557 & 0.8073 & 0.9993 & 0.2022 & 0.4330 & 0.0308 & 0.9536 & 0.9995 & 0.0515 \\
Qwen3-VL-30B-Instruct~\cite{bai2025qwen3} & 0.4215 & 0.5908 & 0.9774 & 0.4011 & 0.1841 & 0.7137 & 0.9701 & 0.2388 & 0.6894 & 0.3661 & 0.7137 & 0.9701 & 0.2388 & 0.5694 & 0.1886 & 0.8276 & 0.9966 & 0.1722 \\
LLaVA-OV-7B~\cite{li2024llava} & 0.0537 & 0.7861 & \underline{0.7930} & 0.1332 & 0.0258 & 0.8339 & \textbf{0.8398} & 0.0838 & 0.3854 & 0.0000 & 1.0000 & 1.0000 & 0.0027 & 0.3367 & 0.0000 & 1.0000 & 1.0000 & 0.0073 \\
LLaVA-OV-72B~\cite{li2024llava} & 0.5149 & 0.6541 & 0.9926 & 0.3625 & 0.2816 & 0.7280 & 0.9703 & 0.2547 & 0.5374 & 0.0683 & 0.8744 & 0.9622 & 0.1024 & 0.3462 & 0.0078 & 0.9869 & 0.9963 & 0.0169 \\
MiniCPM-V-2.6~\cite{yao2024minicpm} & 0.0000 & 1.0000 & 1.0000 & 0.0027 & 0.0000 & 1.0000 & 1.0000 & 0.0073 & 0.3827 & 0.0000 & 1.0000 & 1.0000 & 0.0027 & 0.3377 & 0.0000 & 1.0000 & 1.0000 & 0.0073 \\
MiMo-VL-7B-SFT~\cite{yue2025mimo} & 0.3641 & 0.6317 & 0.8847 & 0.3326 & 0.1569 & 0.8092 & 0.9530 & 0.1620 & 0.5650 & 0.2280 & \underline{0.6539} & \textbf{0.8739} & 0.2914 & 0.3731 & 0.0505 & 0.8866 & \underline{0.9302} & 0.0763 \\
% InternLM-XComposer2.5 & 0.0011 & 0.9986 & 0.9993 & 0.0040 & 0.0005 & 0.9991 & 0.9995 & 0.0079 & 0.3797 & 0.0000 & 1.0000 & 1.0000 & 0.0027 & 0.3318 & 0.0000 & 1.0000 & 1.0000 & 0.0073 \\
% mPLUG-Owl3-7B & 0.1646 & 0.7123 & 0.8402 & 0.2891 & 0.0514 & 0.8428 & 0.8996 & 0.0969 & 0.1386 & 0.0310 & 0.8997 & \underline{0.9171} & 0.0657 & 0.1397 & 0.0130 & 0.9424 & 0.9484 & 0.0358 \\
Idefics2-8B~\cite{laurenccon2024matters} & 0.1667 & 0.6279 & \textbf{0.7653} & 0.2729 & 0.0211 & 0.8827 & 0.8959 & 0.0643 & 0.3870 & 0.0004 & 0.9987 & 0.9987 & 0.0036 & 0.3292 & 0.0001 & 0.9998 & 1.0000 & 0.0074 \\
Mantis-8B~\cite{jiang2024mantis} & 0.2069 & 0.5810 & 0.8146 & 0.3242 & 0.1003 & 0.7227 & 0.8813 & 0.1864 & 0.1282 & 0.0045 & 0.9917 & 0.9980 & 0.0091 & 0.0474 & 0.0000 & 1.0000 & 1.0000 & 0.0073 \\
% Phi-3.5 & 0.1573 & 0.8079 & 0.9359 & 0.1645 & 0.0426 & 0.9019 & 0.9525 & 0.0727 & 0.3964 & 0.0038 & 0.9941 & 0.9980 & 0.0077 & 0.3306 & 0.0006 & 0.9992 & 1.0000 & 0.0080 \\
Phi-4-multimodal-instruct~\cite{abdin2024phi} & 0.0845 & 0.8243 & 0.8847 & 0.1271 & 0.0133 & 0.9558 & 0.9685 & 0.0326 & 0.4001 & 0.0119 & 0.9834 & 0.9966 & 0.0171 & 0.3230 & 0.0010 & 0.9984 & 0.9995 & 0.0087 \\
%\multicolumn{19}{l}{\textit{Proprietary Models}}\\
\midrule
\multicolumn{19}{l}{\textit{Proprietary Models}} \\
\midrule
GPT-5~\cite{gpt5} & 0.4387 & 0.6510 & 0.9825 & 0.3524 & \underline{0.3319} & 0.6586 & 0.9671 & 0.3217 & 0.6573 & 0.2714 & 0.6982 & 0.9651 & 0.2919 & \textbf{0.6312} & 0.1296 & 0.8259 & 0.9786 & 0.1580 \\
Gemini 2.5-Pro~\cite{comanici2025gemini} & \underline{0.5511} & \underline{0.5426} & 0.9791 & \textbf{0.4618} & 0.3023 & \textbf{0.5300} & \underline{0.8717} & \underline{0.3822} & \textbf{0.7311} & \textbf{0.4208} & \textbf{0.5571} & \underline{0.9332} & \textbf{0.4258} & 0.5984 & 0.1857 & 0.7536 & 0.9311 & 0.2133 \\
Claude Sonnet 4.5~\cite{Claude} & \textbf{0.6410} & 0.6241 & 0.9953 & 0.4015 & \textbf{0.5437} & 0.6085 & 0.9922 & \textbf{0.3997} & 0.6240 & 0.3988 & 0.7235 & 0.9980 & 0.2908 & 0.4967 & \underline{0.2036} & 0.8362 & 0.9956 & 0.1696 \\
\midrule
FakeShield~\cite{xu2025fakeshield} & 0.1352 & 0.8315 & 0.9393 & 0.1488 & -- & -- & -- & -- & 0.4045 & 0.0254 & 0.9752 & 0.9974 & 0.0307 & -- & -- & -- & -- & -- \\
FakeVLM~\cite{wen2025spot} & 0.3595 & 0.7792 & 0.9973 & 0.2361 & -- & -- & -- & -- & 0.4736 & 0.0062 & 0.9954 & 1.0000 & 0.0048 & -- & -- & -- & -- & -- \\
\bottomrule
\end{tabular}}}
\begin{tablenotes}[para,flushleft]
\scriptsize
\parbox{\textwidth}{
\textbf{Notes.} -- indicates unsupported modality.}
\end{tablenotes}
\end{table*}

\section{Experiments}
\label{sec:exp}

%\subsection{Evaluation Setup}
%\noindent\textbf{Evaluation models and modalities.} \hky{We benchmark $22$ MLLMs ($19$ open-source, $3$ proprietary) across image, video, and audio modalities. For visual tasks, open-source models include InternVL2\_5/3\_5~\cite{chen2024expanding,wang2025internvl3}, Qwen3-Omni/VL~\cite{xu2025qwen3,bai2025qwen3}, LLaVA-OV~\cite{li2024llava}, MiniCPM-V~\cite{yao2024minicpm}, MiMo-VL~\cite{yue2025mimo}, Idefics2~\cite{laurenccon2024matters}, Mantis~\cite{jiang2024mantis}, Phi-4~\cite{abdin2024phi}, and the forensic-focused FakeShield~\cite{xu2025fakeshield}. These are compared against proprietary baselines: GPT-5~\cite{gpt5}, Gemini 2.5-Pro~\cite{comanici2025gemini}, and Claude Sonnet 4.5~\cite{Claude}. Audio performance is evaluated using Qwen2-Audio~\cite{chu2024qwen2}, Qwen3-Omni, Phi, Audio-Flamingo-3~\cite{ghosh2025audio}, and SALMONN-7B~\cite{tang2024salmonn}, with Gemini 2.5-Pro serving as the proprietary reference.}
%\noindent\textbf{Experimental protocol.} \hky{All evaluations are conducted in a zero-shot setting without task-specific fine-tuning. Each query consists of a text prompt paired with the corresponding image, video, or audio input. For video tasks, we utilize 16-frame sampling where configurable. Otherwise, we adhere to the model’s default sampling policy.}
Due to page limits, the evaluation setup and audio modality results are provided in the supplementary material. In this section, we primarily present results on the visual modality.

%\subsection{Benchmarking Visual Interpretability}
%\noindent\hky{\textbf{Perceptual Evaluation of Visual Artifacts.} }
%Model performance in artifact perception exhibits heterogeneous behavior across \texttt{<TFQ>} and \texttt{<MCQ>} in~\cref{tab:perception}. GPT-5, Qwen3-VL-30B-Instruct, Gemini 2.5-Pro, and LLaVA-OV-72B emerge as the most robust performers, consistently achieving top-tier rankings across both formats, manifesting their stable perception capabilities. In contrast, several models display pronounced discrepancies between the two metrics. For example, Claude Sonnet 4.5 ranks $14$th in \texttt{<TFQ>} but achieves $1$st place in \texttt{<MCQ>}, suggesting that its strength lies in comparative reasoning rather than the absolute identification of individual synthetic artifacts. Furthermore, Qwen3-VL-30B-Instruct and LLaVA-OV-72B demonstrate relatively balanced and competitive performance across both \texttt{<TFQ>} and \texttt{<MCQ>}, outperforming their smaller or earlier counterparts. Although they do not necessarily yield the highest absolute accuracy, the general trend indicates that stronger visual encoders and more advanced vision–language backbones exhibit greater perceptual sensitivity to subtle synthetic irregularities. Despite the advances, the results in~\cref{tab:perception} highlight that robust DeepFake perception remains far from solved. Even leading models only achieve moderate gains over random choice, leaving substantial room for improvement. 
\noindent \textbf{Evaluation of Perception.}
We begin by assessing the perception ability using the \texttt{<TFQ>} and \texttt{<MCQ>} subsets constructed on manipulated samples only, as summarized in~\cref{tab:perception}. These two test sets target complementary aspects of perceptual capability: \texttt{<TFQ>} mainly probes whether a model can reliably verify the presence or absence of a single artifact or location cue, while \texttt{<MCQ>} requires selecting one or more correct options among several plausible candidates and an explicit “none of the above” choice, which reduces the chance of answering by relying solely on option priors. Across both settings, GPT-5 and Gemini 2.5-Pro generally outperform open-source systems, revealing a clear gap in low-level and mid-level DeepFake perception between closed and open models. 

A closer comparison between \texttt{<TFQ>} and \texttt{<MCQ>} reveals that they stress different weaknesses. Claude Sonnet 4.5, for example, achieves the strongest performance on \texttt{<MCQ>} but exhibits a noticeable drop on \texttt{<TFQ>}, suggesting that it can effectively exploit the richer contextual cues and answer structure in multi-choice questions, yet struggles more when forced to make isolated binary judgments without distractor options. In contrast, Qwen3-VL-30B-Instruct and LLaVA-OV-72B achieve relatively balanced and competitive results across both subsets, indicating that stronger visual encoders and larger vision-language backbones do translate into better DeepFake perception, although their absolute accuracy lags behind the best system. 

Overall, these results reveal a clear performance gap between proprietary and open-source MLLMs on both \texttt{<TFQ>} and \texttt{<MCQ>}, and show that robust DeepFake perception is still far from solved. Even the strongest systems only moderately outperform random choice in several settings, indicating substantial headroom for improvement. To pinpoint where current MLLMs actually struggle, we analyze performance across individual artifact types in~\cref{sec:discussion} (RQ1).

\noindent\textbf{Interpretable Detection, Perception and Hallucination Robustness.}
\cref{tab:open-ended} reports benchmarking results on two types of \texttt{<OEQ>}. 
For Type-A \texttt{<OEQ>}, where the input is known to be fake, GPT-5, Gemini 2.5-Pro, Claude Sonnet 4.5, and LLaVa-OV-72B can effectively explain potential artifacts, as indicated by higher \textit{Cover}. On the other hand, \textit{CHAIR} and \textit{Hal} scores are generally high, indicating that hallucinations remain widespread in most model outputs. Overall, $F^{0.5}$-score provides a single weighted indicator that jointly accounts for \textit{Cover} and \textit{CHAIR}, and is suitable for holistic evaluation of interpretability and hallucination.

For Type-B \texttt{<OEQ>}, models must both classify real/fake and provide an explanation. For image modality, Gemini 2.5-Pro and Qwen3-Omni-30B-A3B-Instruct achieve strong \textit{Acc.} and higher \textit{Cover} than others, reflecting stronger explanatory ability. Nonetheless, Qwen3-Omni-30B-A3B-Instruct exhibits pronounce hallucinations, as suggested by its \textit{CHAIR} and \textit{Hal}. For video modality, the drops in \textit{Acc.} for most models reflect increased task difficulty, while the roughly halved \textit{Cover} further highlights the challenge of explaining video DeepFakes.

Our evaluation framework also uncovers behaviors that were previously difficult to characterize, turning qualitative interpretability into quantifiable insights. For example, InternVL2\_5-8B exhibits both low \textit{Cover} (worse) and low \textit{CHAIR} (better). A closer inspection of its predictions shows that the model consistently identifies a small set of artifacts, but the range of artifact types it can detect is notably limited.
In contrast, Claude Sonnet 4.5 attains high \textit{Cover} (better) but relatively high \textit{CHAIR} (worse). Our statistics further show that its average response length is roughly twice that of other models, indicating a stronger tendency toward hallucinated or over-elaborate explanations. This aligns with our earlier observation that, although Claude Sonnet 4.5 demonstrates strong perceptual ability, it still exhibits a pronounced performance gap between \texttt{<TFQ>} and \texttt{<MCQ>}.

\section{Insights and Discussions}
\label{sec:discussion}

%\noindent {\textbf{RQ1. What are the relative difficulties and bottlenecks when detecting quality versus semantic artifacts?}} \hky{For RQ1, we analyze artifact-specific accuracies within the \texttt{<TFQ>} set, as illustrated in~\cref{fig:rad_art}. The results reveal that low-level quality artifacts, such as blockiness, banding, and flicker, are detected with relatively high accuracy. In contrast, semantic artifacts requiring complex physical reasoning, including anatomical inconsistencies and unnatural gazing or blinking, consistently yield lower mean accuracies across all evaluated models. This disparity indicates that while current MLLMs are proficient at identifying local quality defects, semantic reasoning remains the primary bottleneck for achieving robust DeepFake perception.}

\begin{figure}[!t]
  \centering
  \includegraphics[width=\linewidth]{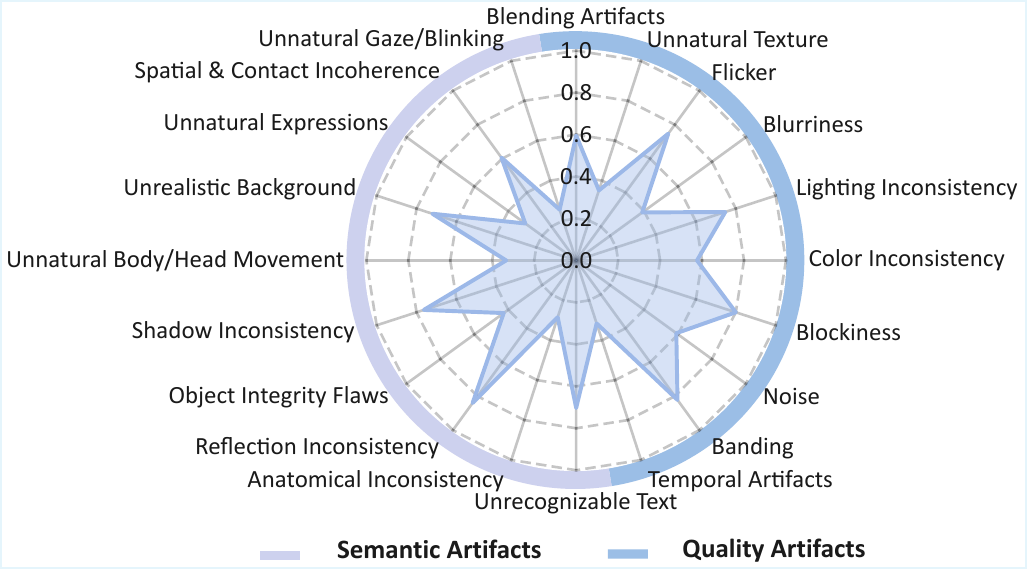} 
  \caption{\pp{Radar chart of accuracy of semantic artifacts and quality artifacts in \texttt{<TFQ>}.}}
  \label{fig:rad_art}
\end{figure}

\noindent \textbf{RQ1. What are the relative difficulties and bottlenecks when detecting quality versus semantic artifacts?}

To address RQ1, we analyze artifact-wise accuracies on the \texttt{<TFQ>} set, as summarized in ~\cref{fig:rad_art}. ~\cref{fig:rad_art} shows the mean accuracy for each artifact type, computed over models with non-zero performance, and reveals that several quality artifacts (e.g., blockiness, banding, reflection inconsistency) can already be detected with relatively high accuracy, even though the overall \texttt{<TFQ>} scores remain moderate.
In contrast, semantic artifacts that require physical or social reasoning (e.g., anatomical inconsistencies, abnormal motion, background–subject incoherence) are consistently much harder, with substantially lower mean accuracies across models. Thus, current MLLMs find local quality artifacts comparatively easier, while semantic artifacts remain the main bottleneck for robust DeepFake perception.

%\noindent {\textbf{RQ2. Do localization-oriented questions truly enhance the model's ability to ``look at the right place''?}} \hky{To quantify the efficacy of localization hints, we define \textit{Benefit} and \textit{Cost} as the percentages of questions where the hint respectively corrects an initial error or induces a new one. Their difference, \textit{Net Benefit}, serves as the primary indicator of genuine performance gain from spatial guidance.}
%\hky{As summarized in~\cref{tab:rq2_benefit_cost}, localization hints generally yield a positive \textit{Net Benefit}, though gains vary by architecture. InternVL2\_5-8B and Claude Sonnet 4.5 achieve peak efficiency ($2.53$\% and $2.47$\% \textit{Net Benefit}), demonstrating an effective ability to leverage spatial cues. Conversely, Gemini 2.5-Pro and Qwen3-VL-30B-Instruct exhibit negative \textit{Net Benefit} ($-0.30$\% and $-0.32$\%), suggesting that for certain high-capacity architectures, external hints may introduce disruptive noise. This non-universal efficacy underscores a persistent architectural gap in reconciling external spatial grounding with internal visual representations.}
\noindent \textbf{RQ2. Do localization-oriented questions truly enhance the model's ability to ``look at the right place''?}

To assess the impact of location hints on model performance, we define two metrics: \textit{Benefit}, the percentage of questions a model answers incorrectly without a location hint but correctly with one; and \textit{Cost}, the percentage of questions a model answers correctly without a hint but incorrectly with one. These metrics highlight model-dependent effects, where hints often yield small gains but substantial losses in performance. Details and the results table are provided in the supplementary material.

A few models demonstrate clear net benefits, leveraging hints effectively with low disruption. For instance, InternVL2\_5-8B and Claude Sonnet 4.5 show modest Benefits with minimal Costs, as do larger variants like InternVL2\_5-26B and 38B. Conversely, some models suffer more harm than help, such as MiniCPM-V-2.6, where Costs far exceed Benefits. Others display high instability, with Benefits nearly matched by Costs, as seen in InternVL3\_5-8B, Qwen3-VL-8B-Instruct, and GPT-5, suggesting unreliable improvements rather than consistent gains.

Overall, localization hints do not reliably improve models' spatial focus. Only select models, like InternVL2\_5-8B and Claude Sonnet 4.5, gain meaningfully with little downside. For most, including strong performers like Gemini 2.5-pro and GPT-5, hints introduce distractions, resulting in limited benefits, instability, or outright setbacks. This reveals difficulties in combining spatial cues with visual tasks. 

%\noindent \textbf{RQ3. How does the tripartite coupling of perception, detection, and hallucination characterize failure modes in MLLM-based DeepFake detection?} \hky{Evaluation across $22$ MLLMs via \benchname~reveals that superior perceptual performance on \texttt{<TFQ>}, \texttt{<MCQ>}, and Type-A \texttt{<OEQ>} does not reliably translate to Type-B detection accuracy. Models with comparable detection scores exhibit significant variance in explanatory coverage and hallucination severity, indicating only moderate coupling between perception and detection. We identify two systematic failure patterns: (1) correct fine-grained perception followed by misclassifications, and (2) comprehensive explanations undermined by hallucinated artifacts. This suggests the perception-to-detection pipeline is frequently disrupted by either evidentiary gaps or hallucination-driven reasoning errors.}
%\hky{From RQ1–RQ3, we conclude that MLLM-based DeepFake detection is inherently three-dimensional, governed by the interaction of perception, detection, and hallucination. While accurate perception is foundational, it is insufficient without robust hallucination mitigation. Reliable detection requires models to both accurately identify existing artifacts and avoid ``perceiving'' non-existent ones.}
\noindent \textbf{RQ3.How are perception, detection, and hallucination coupled in MLLM-based DeepFake detectors, and what failure patterns emerge from this three-dimensional interaction?}

TriDF reveals that strong perceptual performance on \texttt{<TFQ>}, \texttt{<MCQ>}, and Type-A \texttt{<OEQ>} does not reliably translate into Type-B \texttt{<OEQ>} detection accuracy. Models with similar detection scores can differ substantially in explanatory coverage (Cover) and hallucination severity (CHAIR, Hal, $F^{0.5}$), indicating only moderate coupling between perception and detection and a partly independent effect of hallucination. We observe systematic failures where models correctly identify fine-grained artifacts in Type-A \texttt{<OEQ>} yet still misclassify real–fake pairs in Type-B \texttt{<OEQ>}, or produce high-Cover explanations that are contaminated by hallucinated artifacts. These cases show that the perception chain to detection can break either because the model fails to perceive the relevant evidence or because hallucinations distort how this evidence is integrated into a final decision.

Taken together, our findings across RQ1–RQ3 suggest that DeepFake detection in MLLMs is inherently three-dimensional. RQ1 highlights semantic artifacts as a key bottleneck even when many quality artifacts are detectable, and RQ2 shows that localization cues alone do not guarantee that models “look at the right place.” RQ3 further indicates that reliable detection requires both accurate perception and low hallucination: improving DeepFake perception is necessary but not sufficient unless models also avoid “seeing” artifacts that are not there. A more fine-grained three-dimensional analysis (e.g., partial correlations and stratified perception→detection curves under different hallucination regimes) is provided in the supplementary.

\iffalse
\begin{table}[!ht]
\centering
\caption{RQ2. Benefit and Cost of localization hints.}
\label{tab:rq2_benefit_cost}
\renewcommand{\arraystretch}{1.}
\fontsize{6pt}{8pt}\selectfont
{\setlength{\tabcolsep}{0.6mm}
\resizebox{\columnwidth}{!}
{\begin{tabular}{lcccc}
\toprule
\textbf{MLLM} & \textit{Benefit} (\%) & \textit{Cost} (\%) & \textit{Net Benefit} (\%) & \textbf{Rank}\\
\midrule
InternVL2\_5-8B~\cite{chen2024expanding}       & 3.21  & 0.68  & \textbf{\mc{2.53}}  & 1  \\
InternVL2\_5-26B~\cite{chen2024expanding}      & 4.28  & 1.90  & \mc{2.38}  & 4  \\
InternVL2\_5-38B~\cite{chen2024expanding}      & 4.22  & 1.78  & \mc{2.44}  & 3  \\
InternVL3\_5-8B~\cite{wang2025internvl3}       & \underline{10.87} & \underline{10.10} & \mc{0.78}  & 7  \\
InternVL3\_5-38B~\cite{wang2025internvl3}      & 4.93  & 3.30  & \mc{1.63}  & 5  \\
Qwen3-Omni-30B-A3B-Instruct~\cite{xu2025qwen3} & 6.92  & 6.55  & \mc{0.37}  & 9  \\
Qwen3-VL-8B-Instruct~\cite{bai2025qwen3}       & 8.03  & 7.30  & \mc{0.73}  & 8  \\
Qwen3-VL-30B-Instruct~\cite{bai2025qwen3}      & 7.01  & 7.33  & \mc{-0.32} & 11 \\
GPT-5~\cite{gpt5}                              & 6.57  & 5.65  & \mc{0.92}  & 6  \\
Gemini 2.5-Pro~\cite{comanici2025gemini}       & \textbf{11.67} & \textbf{11.97} & \mc{-0.30} & 10 \\
Claude Sonnet 4.5~\cite{Claude}                & 3.17  & 0.70  & \textbf{\mc{2.47}}  & 2  \\
\bottomrule
\end{tabular}}}
\end{table}
\fi

\section{Conclusion}
\label{sec:con}
We present \benchname, a comprehensive benchmark designed to advance interpretable and reliable DeepFake detection. By integrating high-quality synthesized content from a broad spectrum of contemporary generators and providing human-aligned annotations across 16 manipulation types and 3 modalities, \benchname~offers the most extensive resource to date for studying for detection models perceive evidence, make decisions, and articulate their reasoning. Through its three complementary components, \textit{Perception}, \textit{Detection}, and \textit{Hallucination}, our benchmark enables a holistic examination of model behavior that goes beyond traditional accuracy-based evaluation. Our experiments on state-of-the-art multimodal large language models reveal several key findings. Accurate recognition of manipulation cues is essential for strong classification performance, yet unreliable or fabricated explanations can significantly undermine the final decision of a model. The key findings highlight the interdependence of perception, detection, and explanation reliability, and demonstrate the need for evaluation protocols that account for all three.

\clearpage

\section{Acknowledgment}

This work was partially supported by the National Science and Technology Council, Taiwan (Grants: NSTC-112-2628-E-002-033-MY4, NSTC-114-2634-F-002-004, and NSTC-112-2221-E-A49-059-MY3), the Taiwan Centers of Excellence (TCE), and the Center of Data Intelligence: Technologies, Applications, and Systems (Grants: 115L900901/115L900902/115L900903), National Taiwan University, from the Featured Areas Research Center Program within the framework of the Higher Education Sprout Project by the Ministry of Education, Taiwan. This work was also supported by the NVIDIA Academic Grant Program. Access to NVIDIA GPUs and software toolkits enabled us to conduct experiments on large training datasets and inspired new research directions.

{
    \small
    \bibliographystyle{ieeenat_fullname}
    \bibliography{main}
}
\clearpage
\appendix  
\section*{Supplementary Material}
\addcontentsline{toc}{section}{Supplementary Material}
\DoToC
\input{supp_content_C}

\end{document}

%% file: supp_content_C.tex
% \tableofcontents
% \localtableofcontents
\section{DeepFake Tasks in TriDF}
\label{sec:tasks}

\begin{figure*}[!ht]
  \centering
  \includegraphics[width=0.75\linewidth]{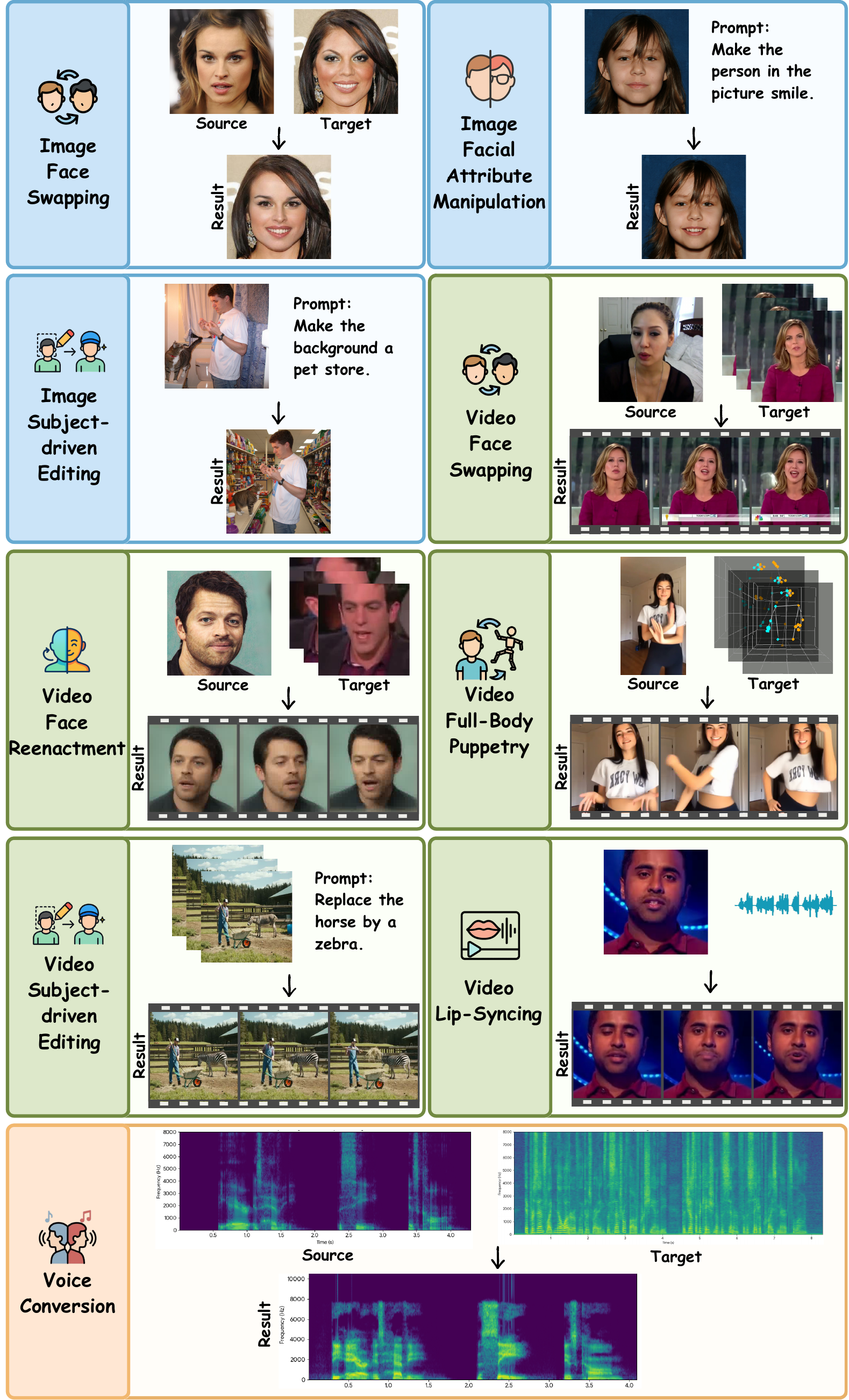}
  \caption{Examples of DeepFakes from Partially Manipulated tasks.}
  \label{fig:partial}
\end{figure*}

\begin{figure*}[!ht]
  \centering
  \includegraphics[width=0.8\linewidth]{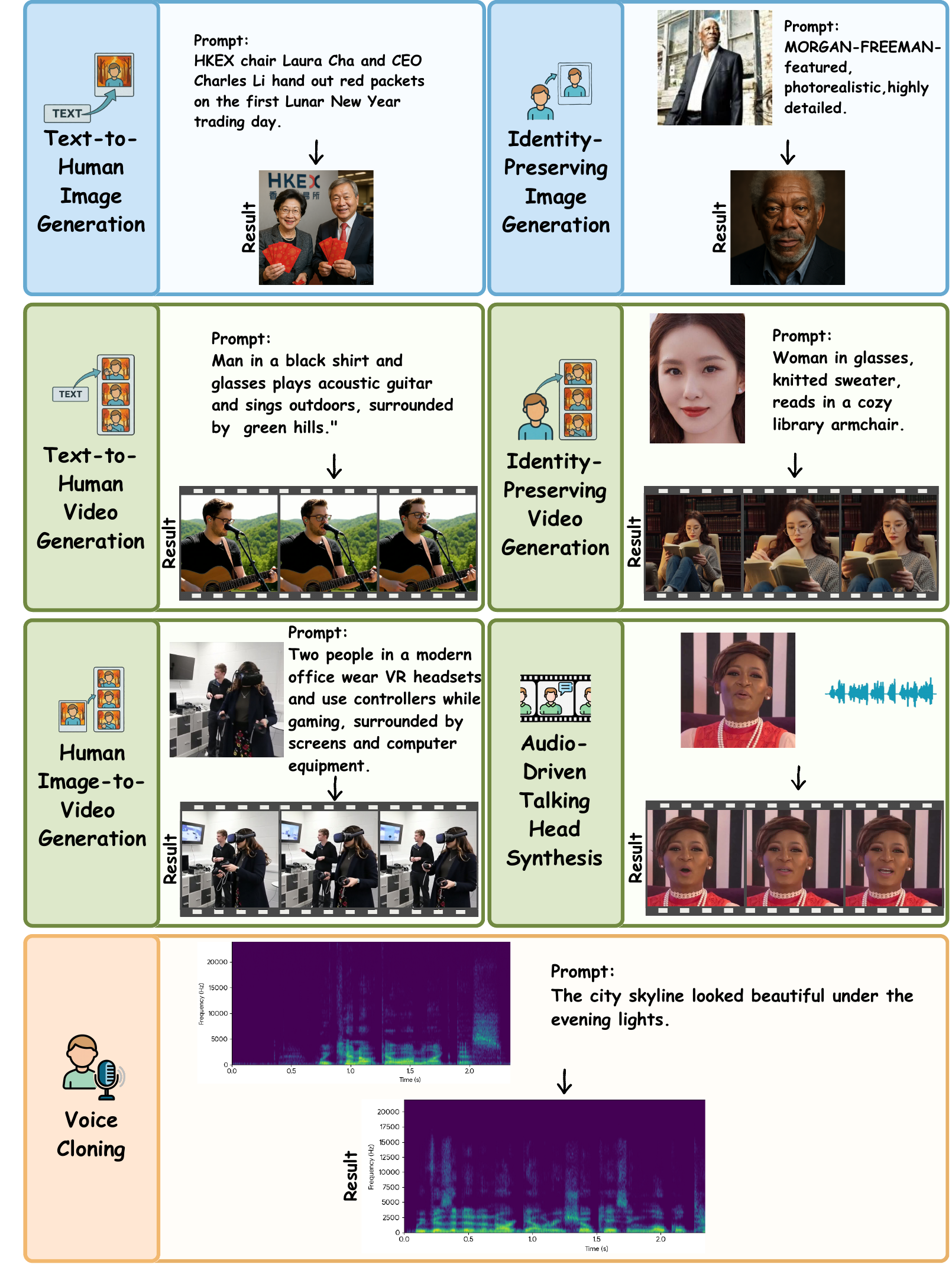}
  \caption{Examples of DeepFakes from Fully Synthesized tasks.}
  \label{fig:fully_synthesis}
\end{figure*}

\hky{DeepFake technologies and synthetic media systems encompass a broad spectrum of manipulation techniques, each targeting distinct aspects of human-centric visual and auditory content. To systematically evaluate this landscape, \benchname\ organizes these techniques into two functional categories: Partially Manipulated, which encompasses methods that alter specific attributes of an existing subject within a scene, and Fully Synthesized, which covers approaches that generate entirely artificial human appearances or voices. Representative qualitative examples for each category are illustrated in~\cref{fig:partial,fig:fully_synthesis}, respectively. In what follows, we formally define each category included in \benchname\ and characterize its distinguishing properties, clarifying how each contributes to the benchmark's comprehensive coverage of the DeepFake detection problem.}
% \hky{DeepFake technologies and synthetic media applications rely on a variety of underlying tasks to alter or generate human-centric content. For the purpose of constructing \benchname, we group the manipulations into two functional categories: Partially Manipulated, which alters an existing person in the scene, and Fully Synthesized, which creates artificial humans or voices without requiring a real subject. The corresponding qualitative samples are provided in~\cref{fig:partial} and~\cref{fig:fully_synthesis}. Below, we outline the categories included in \benchname\ and briefly describe their defining characteristics to clarify how they contribute to the benchmark’s coverage.}

\subsection{Partially Manipulated Tasks}

\noindent \textbf{Image/Video Face Swapping} transfers the identity of a source subject onto a target face while preserving the target's original scene-consistent attributes, including pose, illumination, and expression.

\noindent \textbf{Facial Attribute Manipulation} selectively modifies specific semantic facial attributes, such as age, expression, hair color, or accessories, in a directed and controlled manner, while preserving the subject's core identity.

\noindent \textbf{Lip Synchronization} alters the lip movements of a subject in a video to match a new or substituted audio track, producing the perceptual illusion that the subject is articulating words they did not originally utter.

\noindent \textbf{Face Reenactment} transfers the facial expressions, head pose, and eye gaze of a source subject onto a target subject, effectively compelling the target to replicate the source's performance across a static image or an independent video sequence.

\noindent \textbf{Full-Body Puppetry} extends the face reenactment paradigm to the full human body, transferring the complete skeletal pose and motion dynamics of a source actor onto a target subject, thereby enabling the source to drive the target's movements throughout a video.

\noindent \textbf{Subject-Driven Image/Video Editing} applies targeted manipulations to a specific subject within an image or video, typically guided by textual prompts or reference images (\eg, ``change the person's shirt to red''), while preserving both the subject's identity and the surrounding scene context.

\noindent \textbf{Voice Conversion} transforms a speaker's vocal characteristics to resemble those of a designated target speaker, while strictly preserving the original linguistic content and spoken words.

\subsection{Fully Synthesized Tasks}

\noindent \textbf{Audio-Driven Talking Head Synthesis} generates a fully synthetic video of a human subject in which lip movements, facial expressions, and head pose are produced entirely from scratch and conditioned on an input audio signal, without relying on any real video footage of the subject.

\noindent \textbf{Identity-Preserving Image/Video Generation} synthesizes novel images or videos of a specific individual by learning their identity representation from a limited set of reference photographs, enabling generation of that individual in previously unseen poses, environments, or visual styles.

\noindent \textbf{Text-to-Human Image/Video Generation} involves the synthesis of high-fidelity human images or video sequences conditioned exclusively on textual descriptions. Given a text prompt, generative models map semantic concepts to visually coherent representations without the aid of external visual priors.

\noindent \textbf{Human Image-to-Video Generation} focuses on animating a static reference image into a continuous video sequence, guided by a textual prompt. The objective is to preserve the identity and fine-grained attributes of the source subject while synthesizing realistic motion and temporal dynamics that align with the provided textual instructions.

\noindent \textbf{Voice Cloning} constructs a comprehensive generative model of a specific individual's voice, often from a minimal audio sample, capturing their unique tonal quality, cadence, and vocal style. The resulting model enables arbitrary speech synthesis in the target speaker's voice via text-to-speech generation.

\section{DeepFake Data Generation}
\label{sec:gen}
\noindent \hky{\textbf{Data Acquisition}. We exclusively collect information in accordance with the specific licensing agreements of source websites, avoiding material that is protected against usage for any commercial purposes. The licenses of the existing datasets used in this work are as follows:}
\begin{itemize}
    \item FaceForensics++~\cite{rossler2019faceforensics}: Non-commercial research and educational purposes.
    \item FFHQ~\cite{karras2019style}: Creative Commons BY-NC-SA 4.0
    \item CelebAMaskHQ~\cite{lee2020maskgan}: Non-commercial research and educational purposes.
    \item CelebA-HQ~\cite{karras2018progressive}: Non-commercial research and educational purposes
    \item VGGFace2~\cite{cao2018vggface2}: Unspecified
    \item Emu Edit~\cite{sheynin2024emu}: Creative Commons BY-NC 4.0
    \item GEdit-Bench~\cite{liu2025step1x}: MIT License
    \item ImgEdit~\cite{ye2025imgedit}: Apache license 2.0
    \item OmniContext~\cite{wu2025omnigen2}: Apache License 2.0
    \item MS-COCO~\cite{lin2014microsoft}: Creative Commons BY 4.0
    \item Flickr30k~\cite{plummer2015flickr30k}: Non-commercial research and educational purposes.
    \item LAION-Aesthetics~\cite{schuhmann2022laion}: Creative Commons BY 4.0
    \item VoxCeleb2~\cite{chung2018voxceleb2}: Creative Commons BY-SA 4.0
    \item LRS2~\cite{son2017lip}: Academic Research Purposes.
    \item TalkingHead-1KH~\cite{wang2021one}: Creative Commons BY 3.0
    \item VPBench~\cite{bian2025videopainter}: The CogVideoX License
    \item FiVE-Bench~\cite{li2025five}: Creative Commons BY-NC 4.0
    \item HDTF~\cite{zhang2021flow}: Creative Commons BY 4.0
    \item CelebV-Text~\cite{yu2023celebv}: Non-commercial research purposes only.
    \item Fashion Video~\cite{zablotskaia2019dwnet}: Creative Commons BY-NC 4.0
    \item TED-talks~\cite{siarohin2021motion}: Unspecified
    \item TikTok~\cite{jafarian2022self}: Creative Commons BY-NC 4.0
    \item A2 Bench~\cite{fei2025skyreels}: Apache License 2.0
    \item OpenS2V-Nexus~\cite{yuan2025opens2v}: Apache License 2.0
    \item ConsisID~\cite{yuan2025identity}: Creative Commons BY 4.0
    \item Panda-70M~\cite{chen2024panda}: Non-commercial and research purposes.
    \item HOIGen-1M~\cite{liu2025hoigen}: Apache License 2.0
    \item EMIME~\cite{wester2010emime}: Open Data Commons Attribution License (ODC-By) v1.0
    \item VCTK~\cite{veaux2017cstr}: Creative Commons BY 4.0
    \item LibriTTS~\cite{zen2019libritts}: Creative Commons BY 4.0
    \item LibriSpeech~\cite{panayotov2015librispeech}: Creative Commons BY 4.0
\end{itemize}

\noindent \hky{All datasets released with this work are available under the Creative Commons Attribution-NonCommercial-ShareAlike 4.0 International license (CC BY-NC-SA 4.0). We selected this license to match the terms of several original datasets and to provide our data under the same access conditions.}

\noindent \hky{\textbf{Data Generation}. To ensure comprehensive coverage, we organize our synthesis pipeline into task-oriented sub-domains, as detailed in~\cref{tab:task}.}
% To ensure comprehensive coverage of the DeepFake landscape, we organize our synthesis pipeline into distinct task-oriented sub-domains, as detailed in~\cref{tab:task}.}

\hky{In the image modality, we move beyond traditional Face Swapping to include Subject-driven Editing and Identity-Preserving Generation, utilizing both open-source models, such as PixArt-$\sigma$~\cite{chen2024pixart}, OmniGen2~\cite{wu2025omnigen2}, Step1X-Edit~\cite{liu2025step1x}, SD3~\cite{esser2024scaling}, and Flux~1~\cite{batifol2025flux}, and proprietary generators like Gemini 2.5~\cite{nanoBanana} and GPT‑4o~\cite{gptImage}.}
% \noindent \hky{In the Image modality, we extend beyond traditional Face Swapping and Attribute Manipulation to encompass advanced Subject-driven Editing and Identity-Preserving Generation. This involves a diverse array of state-of-the-art models, ranging from open-source editors like PixArt-$\sigma$~\cite{chen2024pixart}, OmniGen2~\cite{wu2025omnigen2}, Step1X-Edit~\cite{liu2025step1x}, SD3~\cite{esser2024scaling}, and Flux~1~\cite{batifol2025flux}, to proprietary generators such as Gemini 2.5~\cite{nanoBanana} and GPT‑4o~\cite{gptImage}.}

\hky{The video modality represents the most diverse category, addressing the spectrum from facial to full-body synthesis. We include head-centric tasks, such as Face Reenactment and Lip-Syncing (\eg, MuseTalk~\cite{zhang2025musetalk}), alongside complex body-centric tasks like Full-Body Puppetry via Champ~\cite{zhu2024champ} and ControlNeXt~\cite{peng2024controlnext}. Furthermore, we incorporate Human Video Generation utilizing models like LTX-Video~\cite{hacohen2024ltx}, Wan2.2~\cite{wan2025wan}, Phantom~\cite{liu2025phantom}, and HunyuanCustom~\cite{hu2025hunyuancustom}, covering various conditioning inputs such as reference images and pure text.}

\noindent \hky{Finally, for the audio modality, we target both Voice Cloning and Voice Conversion. By gathering open-source solutions like OpenVoice~\cite{qin2023openvoice} and Seed-VC~\cite{liu2024zero} against commercial APIs like ElevenLabs~\cite{elevenlabs}, we capture the current state-of-the-art across varying acoustic environments.}

\noindent \hky{\textbf{Quality Control}. To increase the high fidelity of our generated DeepFakes, we employ specialized metrics for assessing realism and consistency to ensure automatic quality control before starting the annotation process. \textit{Realism metrics}, namely LPIPS~\cite{zhang2018unreasonable}, NIQE~\cite{mittal2012making}, VSFA~\cite{li2019quality}, and NISQA~\cite{mittag2021nisqa}, evaluate whether the content appears natural and is challenging for humans or algorithms to detect as synthetic. In contrast, \textit{consistency metrics}, including ArcFace~\cite{deng2019arcface}, CLIPScore~\cite{hessel2021clipscore}, LSE-C~\cite{prajwal2020lip}, AED\&AKD~\cite{siarohin2019first}, SECS~\cite{liu2024zero}, and ViCLIP~\cite{wang2023internvid}, measure how closely the output aligns with input conditions or control signals, such as retaining facial identity, voice characteristics, or movement synchronization. After applying quality control, we form one-to-one real-fake pairs in each DeepFake task, resulting in a total of over $5$K high-quality pairs spanning three different modalities.}

\begin{table*}[!ht]
    \centering
    \caption{\hky{Overview of DeepFake tasks, representative synthesis methods, and commonly used public datasets across three modalities. For each task, we select three publicly available code repositories to ensure diversity in generation approaches. To maintain fair evaluation and simulate real-world scenarios, only the testing splits of public datasets or datasets not used for training are employed for generation.}}
    \resizebox{0.8\linewidth}{!}{
    \begin{tabular}{p{1.4cm}|p{5.2cm}|p{4.5cm}|p{4cm}}
        \hline 
        \textbf{Modality} & \textbf{Tasks} & \textbf{Synthesis Methods} & \textbf{Public Dataset} \\ \hline 
        % Image
        \multirow{18}{*}{Image}
        & \multirow{3}{*}{Face Swapping}
        & DiffSwap~\cite{zhao2023diffswap} & \multirow{3}{*}{\parbox{4cm}{FaceForensics++~\cite{rossler2019faceforensics}\\FFHQ~\cite{karras2019style}\\CelebAMaskHQ~\cite{lee2020maskgan}}} \\ \cline{3-3}
        &   & BlendFace~\cite{shiohara2023blendface} & \\ \cline{3-3}
        &   & CSCS~\cite{huang2024identity} & \\ \cline{2-4}

        & \multirow{3}{*}{Facial Attribute Manipulation}
        & PREIM3D~\cite{li2023preim3d} & \multirow{3}{*}{\parbox{4cm}{CelebA-HQ~\cite{karras2018progressive}\\VGGFace2~\cite{cao2018vggface2}\\FFHQ~\cite{karras2019style}}} \\ \cline{3-3}
        &   & AdaTrans~\cite{huang2023adaptive} & \\ \cline{3-3}
        &   & StyleGANEX~\cite{yang2023styleganex} & \\ \cline{2-4}

        & \multirow{4}{*}{Subject-driven Image Editing}
        & Mige~\cite{tian2025mige} & \multirow{4}{*}{\parbox{4cm}{Emu Edit~\cite{sheynin2024emu}\\GEdit-Bench~\cite{liu2025step1x}\\ImgEdit~\cite{ye2025imgedit}}} \\ \cline{3-3}
        &   & Step1X-Edit~\cite{liu2025step1x} & \\ \cline{3-3}
        &   & OmniGen2~\cite{wu2025omnigen2} & \\ \cline{3-3}
        &   & Gemini 2.5 Flash Image~\cite{nanoBanana} & \\ \cline{2-4}

        & \multirow{4}{*}{Identity-Preserving Generation}
        & Mige~\cite{tian2025mige} & \multirow{4}{*}{\parbox{4cm}{CelebA-HQ~\cite{karras2018progressive}\\FFHQ~\cite{karras2019style}\\OmniContext~\cite{wu2025omnigen2}}} \\ \cline{3-3}
        &   & UNO~\cite{wu2025less} & \\ \cline{3-3}
        &   & OmniGen2~\cite{wu2025omnigen2} & \\ \cline{3-3}
        &   & Gemini 2.5 Flash Image~\cite{nanoBanana} & \\ \cline{2-4}

        & \multirow{4}{*}{Human Scene Generation}
        & SD3~\cite{esser2024scaling} & \multirow{4}{*}{\parbox{4cm}{MS-COCO~\cite{lin2014microsoft}\\Flickr30k~\cite{plummer2015flickr30k}\\LAION-Aesthetics~\cite{schuhmann2022laion}}} \\ \cline{3-3}
        &   & PixArt-$\sigma$~\cite{chen2024pixart} & \\ \cline{3-3}
        &   & Flux 1.~\cite{batifol2025flux} & \\ \cline{3-3}
        &   & GPT‑4o Image~\cite{gptImage} & \\ \cline{2-4}
        \hline

        % Video
        \multirow{32}{*}{Video}
        
        & \multirow{3}{*}{Face Swapping}
        & HifiFace~\cite{wang2021hififace} & \multirow{3}{*}{\parbox{4cm}{CelebA-HQ~\cite{karras2018progressive}\\VoxCeleb2~\cite{chung2018voxceleb2}\\FaceForensics++~\cite{rossler2019faceforensics}}} \\ \cline{3-3}
        &   & InfoSwap~\cite{gao2021information} & \\ \cline{3-3}
        &   & FaceAdapter~\cite{han2024face} & \\ \cline{2-4}
        
        & \multirow{3}{*}{Face Reenactment}
        & MCNet~\cite{hong2023implicit} & \multirow{3}{*}{\parbox{4cm}{CelebA-HQ~\cite{karras2018progressive}\\VoxCeleb2~\cite{chung2018voxceleb2}\\FaceForensics++~\cite{rossler2019faceforensics}}} \\ \cline{3-3}
        &   & HyperReenact~\cite{bounareli2023hyperreenact} & \\ \cline{3-3}
        &   & LivePortrait~\cite{guo2024liveportrait} & \\ \cline{2-4}
         
        & \multirow{3}{*}{Lip-Syncing}
        & DINet~\cite{zhang2023dinet} & \multirow{3}{*}{\parbox{4cm}{LRS2~\cite{son2017lip}\\VoxCeleb2~\cite{chung2018voxceleb2}\\TalkingHead-1KH~\cite{wang2021one}}} \\ \cline{3-3}
        &   & LatentSync~\cite{li2024latentsync} & \\ \cline{3-3}
        &   & MuseTalk~\cite{zhang2025musetalk} & \\ \cline{2-4}

        & \multirow{3}{*}{Subject-driven Video Editing}
        & VideoPainter~\cite{bian2025videopainter} & \multirow{3}{*}{\parbox{4cm}{VPBench~\cite{bian2025videopainter}\\FiVE-Bench~\cite{li2025five}}} \\ \cline{3-3}
        &   & VACE~\cite{jiang2025vace} & \\ \cline{3-3}
        &   & Wan-Edit~\cite{li2025five} & \\ \cline{2-4}

        & \multirow{4}{*}{Audio-driven Talking-Head Synthesis}
        & SadTalker~\cite{zhang2023sadtalker} & \multirow{4}{*}{\parbox{4cm}{TalkingHead-1KH~\cite{wang2021one}\\HDTF~\cite{zhang2021flow}\\CelebV-Text~\cite{yu2023celebv}}} \\ \cline{3-3}
        &   & AniPortrait~\cite{wei2024aniportrait} & \\ \cline{3-3}
        &   & Hallo2~\cite{cui2025hallo} & \\ \cline{3-3}
        &   & D-ID~\cite{D-ID} & \\ \cline{2-4}
       
        & \multirow{4}{*}{Full-Body Puppetry}
        & Champ~\cite{zhu2024champ} & \multirow{4}{*}{\parbox{4cm}{Fashion Video~\cite{zablotskaia2019dwnet}\\TED-talks~\cite{siarohin2021motion}\\TikTok~\cite{jafarian2022self}}} \\ \cline{3-3}
        &   & MotionEditor~\cite{tu2024motioneditor} & \\ \cline{3-3}
        &   & MagicDance~\cite{chang2024magicpose} & \\ \cline{3-3}
        &   & ControlNeXt~\cite{peng2024controlnext} & \\ \cline{2-4}

        & \multirow{4}{*}{Identity-Preserving Generation}
        & Hunyuancustom~\cite{hu2025hunyuancustom} & \multirow{4}{*}{\parbox{4cm}{A2 Bench~\cite{fei2025skyreels}\\OpenS2V-Nexus~\cite{yuan2025opens2v}\\ConsisID~\cite{yuan2025identity}}} \\ \cline{3-3}
        &   & VACE~\cite{jiang2025vace} & \\ \cline{3-3}
        &   & Phantom~\cite{liu2025phantom} & \\ \cline{3-3}
        &   & Kling~\cite{ding2025kling} & \\ \cline{2-4}

        & \multirow{4}{*}{Human Image-to-Video Generation}
        & LTX-Video~\cite{hacohen2024ltx} & \multirow{4}{*}{\parbox{4cm}{CelebV-Text~\cite{yu2023celebv}\\Panda-70M~\cite{chen2024panda}\\HOIGen-1M~\cite{liu2025hoigen}}} \\ \cline{3-3}
        &   & CogVideoX~\cite{yang2024cogvideox} & \\ \cline{3-3}
        &   & Wan2.2~\cite{wan2025wan} & \\ \cline{3-3}
        &   & Veo3~\cite{veo3} & \\ \cline{2-4}

        & \multirow{4}{*}{Human Scene Generation}
        & LTX-Video~\cite{hacohen2024ltx} & \multirow{4}{*}{\parbox{4cm}{CelebV-Text~\cite{yu2023celebv}\\Panda-70M~\cite{chen2024panda}\\HOIGen-1M~\cite{liu2025hoigen}}} \\ \cline{3-3}
        &   & Pyramid-Flow~\cite{jin2024pyramidal} & \\ \cline{3-3}
        &   & SkyReels-A2~\cite{fei2025skyreels} & \\ \cline{3-3}
        &   & Veo3~\cite{veo3} & \\ \cline{2-4}
        \hline

        % Audio
        \multirow{7}{*}{Audio}
        & \multirow{4}{*}{Voice Cloning}
        & XTTS~\cite{coqui_xtts} & \multirow{4}{*}{\parbox{4cm}{EMIME~\cite{wester2010emime}\\VCTK~\cite{veaux2017cstr}\\LibriTTS~\cite{zen2019libritts}}} \\ \cline{3-3}
        &   & OpenVoice~\cite{qin2023openvoice} & \\ \cline{3-3}
        &   & CosyVoice 2.0~\cite{du2024cosyvoice} & \\ \cline{3-3}
        &   & ElevenLabs~\cite{elevenlabs} & \\ \cline{2-4}
        
        & \multirow{3}{*}{Voice Conversion}
        & SpeechT5\_VC~\cite{ao2022speecht5} & \multirow{3}{*}{\parbox{4cm}{LibriSpeech~\cite{panayotov2015librispeech}\\VCTK~\cite{veaux2017cstr}\\LibriTTS~\cite{zen2019libritts}}} \\ \cline{3-3}
        &   & Seed-VC~\cite{liu2024zero} & \\ \cline{3-3}
        &   & Diff-HierVC~\cite{choi2023diff} & \\ \cline{2-4}
        \hline
    \end{tabular}
    }
    \label{tab:task}
\end{table*}

\section{Taxonomy of DeepFake Artifacts}
\label{sec:art}
\noindent \hky{To systematically categorize the artifacts present in DeepFake media, we divide the artifacts into two distinct classes based on the level of analysis required for detection. \cref{tab:quality_artifacts} outlines \textit{Quality Artifacts}, which encompass low-level signal distortions and compression errors that are often detectable through traditional image or audio processing techniques. In contrast, \cref{tab:semantic_artifacts} details \textit{Semantic Artifacts}, which represent high-level logical inconsistencies, \eg, violations of physics or anatomy, that require contextual understanding to identify.}

\begin{table*}[ht]
    \centering
    \caption{Quality Artifacts: Localized signal errors detectable by traditional processing methods.}
    \resizebox{0.9\linewidth}{!}{
    \begin{tabular}{@{} l l p{10cm} @{}}
        \toprule
        \textbf{Domain} & \textbf{Artifact} & \textbf{Definition} \\
        \midrule
        \textbf{Visual Signal} & Blurriness & The loss of sharpness and fine detail, making the image appear out of focus. \\
        & Blockiness & Visible square or rectangular patterns on the screen. \\
        & Noise & Random, fine speckles or a sandy texture across the image. \\
        & Banding & Distinct, abrupt steps or bands in areas that should have a smooth color gradient. \\
        & Color Inconsistency & Colors appear unnatural, with excessive saturation or vibrancy. \\
        & Blending Artifacts & Visible boundaries where elements should merge smoothly. \\
        & Lighting Inconsistency & Illumination that does not agree across the scene. \\
        & Unnatural Texture & The surface is overly smooth, missing natural irregularities. \\ \hline
        \addlinespace
        \textbf{Temporal} & Temporal Artifacts & Inconsistencies across frames that break motion continuity. \\
        & Flicker & Noticeable and often rapid variation in the overall brightness. \\ \hline
        \addlinespace
        \textbf{Audio Signal} & Clipping & Harsh, fuzzy, or crackling sound when audio is too loud. \\
        & Hiss & High-frequency static noise (e.g., ``shhhh'' sound). \\
        & Buzz & Low-frequency tone, typically caused by electrical interference. \\
        & Pops & Abrupt, short, and sharp sounds that interrupt the audio. \\
        \bottomrule
    \end{tabular}
    }
    \label{tab:quality_artifacts}
\end{table*}

\begin{table*}[ht]
    \centering
    \caption{Semantic Artifacts: High-level inconsistencies requiring contextual understanding. (Env. = Environment; Lang. = Language)}
    \resizebox{0.9\linewidth}{!}{
    \begin{tabular}{@{} l l p{9cm} @{}}
        \toprule
        \textbf{Context} & \textbf{Artifact} & \textbf{Definition} \\
        \midrule
        \textbf{Physics \& Env.} & Reflection Inconsistency & Reflections do not match the subject, lighting, or scene geometry. \\
        & Shadow Inconsistency & Shadows do not match the subject, lighting, or scene geometry. \\
        & Spatial Incoherence & Objects or people fail to make contact with surfaces or each other. \\
        & Unrealistic Background & Background lacks plausible detail, perspective, or depth. \\ \hline
        \addlinespace
        \textbf{Human Biology} & Anatomical Inconsistency & Human anatomy is implausible (e.g., distorted limbs). \\
        & Unnatural Expressions & Facial expressions do not align with emotion or context. \\
        & Unnatural Gaze & Eye direction or blink behavior appears robotic. \\
        & Unnatural Movement & Motion lacks physical plausibility. \\ \hline
        \addlinespace
        \textbf{Objects \& Lang.} & Object Integrity Flaws & The object is incomplete, broken, or internally inconsistent. \\
        & Unrecognizable Text & Text is unrecognizable, incomplete, broken, or distorted. \\
        & Unnatural Prosody & Speech sounds robotic, monotonous, or flat. \\
        \bottomrule
    \end{tabular}
    }
    \label{tab:semantic_artifacts}
\end{table*}

\begin{figure}[!ht]
  \centering
  \includegraphics[width=\linewidth]{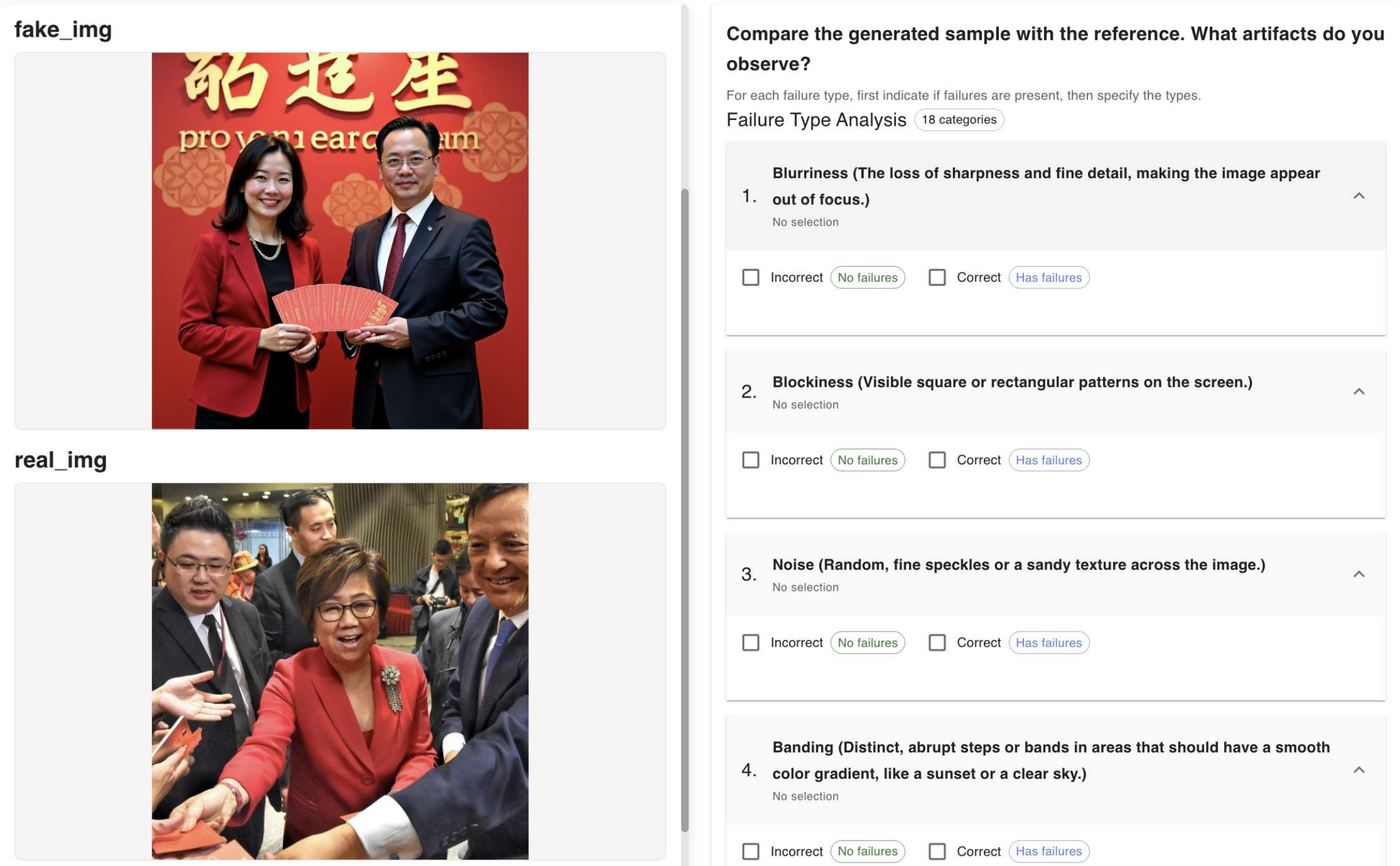}
  \caption{Graphic User Interface of Annotation Platform. It displays paired real and DeepFake samples stacked vertically to facilitate fine-grained comparison and structured artifact labeling for reliable annotation results.}
  \label{fig:anno}
\end{figure}

\section{Annotation Platform}
\label{sec:anno}
\noindent \hky{To implement the unified taxonomy at scale, we have developed a dedicated annotation platform optimized for hierarchical annotation. The annotation process is fully manual, prioritizing accuracy and reliability over automation. In light of the 59\% accuracy ceiling observed with GPT-4o~\cite{gpt4o} on DeepFake detection, reported by LOKI~\cite{ye2025loki}, we have intentionally excluded AI-assisted pre-annotation. We recruited more than 50 annotators. Each generated DeepFake sample is assigned to at least three annotators, and consensus is reached through majority voting. A key feature of our platform, illustrated in~\cref{fig:anno}, is the top-down layout for comparing real and fake media pairs, each matched in a strict one-to-one correspondence. This layout enables annotators to systematically compare manipulated samples with their authentic counterparts, facilitating the precise identification of both \textit{Quality} and \textit{Semantic Artifacts}. To accelerate the annotation process and alleviate the burden of typing complete sentences to describe artifacts found in the generated DeepFake samples, we designed an interface that supports a structured checklist in a multiple-choice style, allowing annotators to assign taxonomy-based labels at multiple levels of granularity with ease and efficiency.}

\begin{figure}[!ht]
  \centering
  \includegraphics[width=\linewidth]{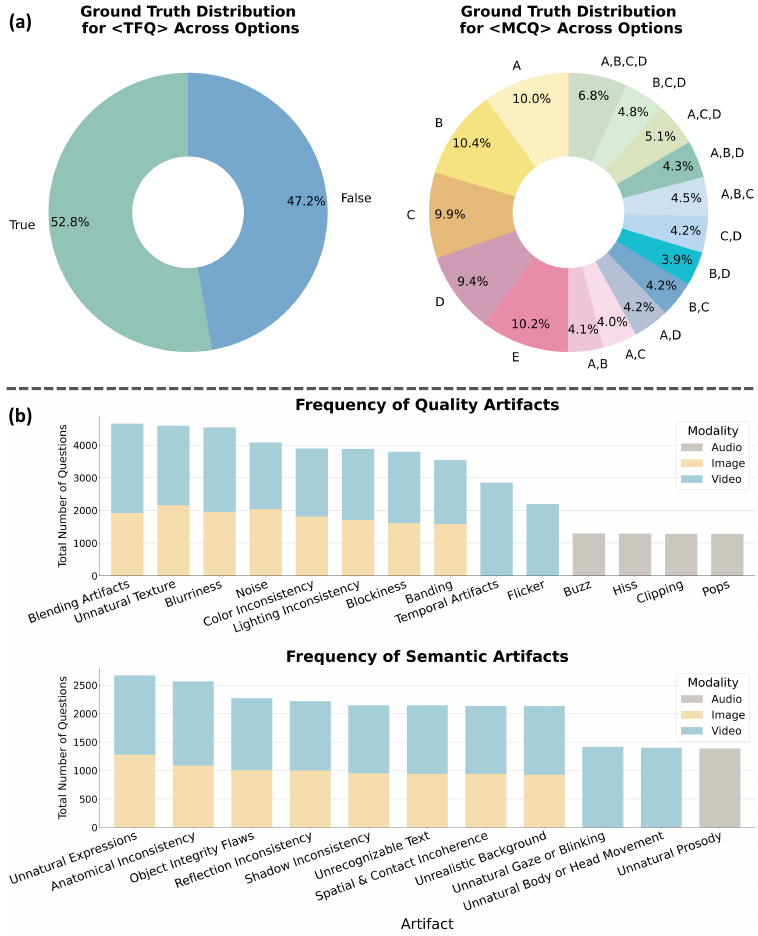}
  \caption{\textbf{Statistics of~\benchname}. (a) The distribution of ground truth options for \texttt{<TFQ>} and \texttt{<MCQ>}. (b) The frequency of quality artifacts and semantic artifacts.}
  \label{fig:stats}
\end{figure}

\section{Distribution of Ground Truth Options}
\label{sec:dis_gt}
\noindent \hky{As illustrated in~\cref{fig:stats}, we adopt the approach from~\cite{loginova2025addressing,zheng2023large} to ensure that the ground truth options, \eg, true-false or multiple-choice options, are distributed as evenly as possible. This step helps alleviate the well-known ``selection bias'' issues in MLLMs~\cite{zheng2023large,min2022rethinking}, where they often favor specific option labels as answers.}

\section{Benchmark Statistics}
\label{sec:stats}
\noindent \hky{\textbf{Comparison with Existing Benchmarks}. As shown in~Tab. 1 in the main paper, we compare our proposed~\benchname~with existing benchmarks~\cite{zhang2024common,li2024fakebench,zhou2025aigi,wang2025forensics,ye2025loki} for DeepFake detection across several key dimensions, including the size of testing sets, the number of generators, the types of DeepFakes, the data modalities, and the evaluation metrics. Notably, \benchname~distinguishes itself with the largest number of questions ($65$K), generators ($51$), and DeepFake types ($16$), spanning three modalities, image, video, and audio, surpassing prior works that often focus on limited generators or types of DeepFake. This extensive collection of generators is a key advantage, providing a far more rigorous test of a detector's robustness and generalization capabilities. It ensures that models are evaluated against a diverse spectrum of generation artifacts, rather than overfitting to the signatures of a few common tools. Crucially, this diversity enables \benchname~ to simulate real-world ``in-the-wild'' scenarios by assessing performance against the latest generation models, including state-of-the-art methods such as PixArt-$\sigma$~\cite{chen2024pixart}, OmniGen2~\cite{wu2025omnigen2}, Step1X-Edit~\cite{liu2025step1x}, Flux 1.~\cite{batifol2025flux}, SD3~\cite{esser2024scaling}, Gemini 2.5 Flash Image~\cite{nanoBanana}, GPT‑4o Image~\cite{gptImage}, Hunyuancustom~\cite{hu2025hunyuancustom}, LTX-Video~\cite{hacohen2024ltx}, Wan2.2~\cite{wan2025wan}, and Veo3~\cite{veo3}. Unlike existing benchmarks, \benchname~features a comprehensive suite of metrics to quantify the interpretability of DeepFake detection, including Accuracy and \textit{Cover} metrics. It also evaluates the perception abilities and hallucination tendencies of MLLMs through strict real-fake pairs, which enable side-by-side comparisons and allow annotators to assign taxonomy-based labels at multiple levels of granularity. This approach provides a more nuanced and robust assessment of model performance in real-world DeepFake scenarios. In designing~\benchname, we deliberately avoid using LLM-as-a-judge approaches. As discussed in~\cite{li2025generation}, employing LLMs as judges inherently introduces biases that can compromise the fairness and reliability of evaluations. Furthermore, LLM judges are susceptible to adversarial attacks, such as prompt injection, thereby raising significant concerns about their reliability in high-stakes scenarios, including DeepFake detection.}

\noindent \hky{\textbf{Statistics}. \benchname~is a meticulously curated benchmark designed to comprehensively evaluate DeepFake detection. It consists of $65$K questions that span $16$ DeepFake techniques, including modern methods like GANs, SD, and DiT. The benchmark's scope is intentionally broad, covering $3$ distinct modalities (image, video, and audio) and multiple types of forgeries, from partially manipulated content to fully synthetic media. To ensure a thorough evaluation of interpretability in DeepFake detection, perception abilities, and hallucination tendencies in MLLMs, the questions are distributed across $23$K \texttt{<TFQ>}, $24$K \texttt{<MCQ>}, and $18$K \texttt{<OEQ>}. This significant diversity challenges MLLMs, requiring them to demonstrate robust generalization and a more comprehensive capacity for identifying different forms of DeepFakes.}

\section{Templates}
\label{sec:temp}
\subsection{Templates for Benchmark Construction}
\label{sec:temp_benchmark}

\begin{figure}[t]
  \centering
  \includegraphics[width=\linewidth]{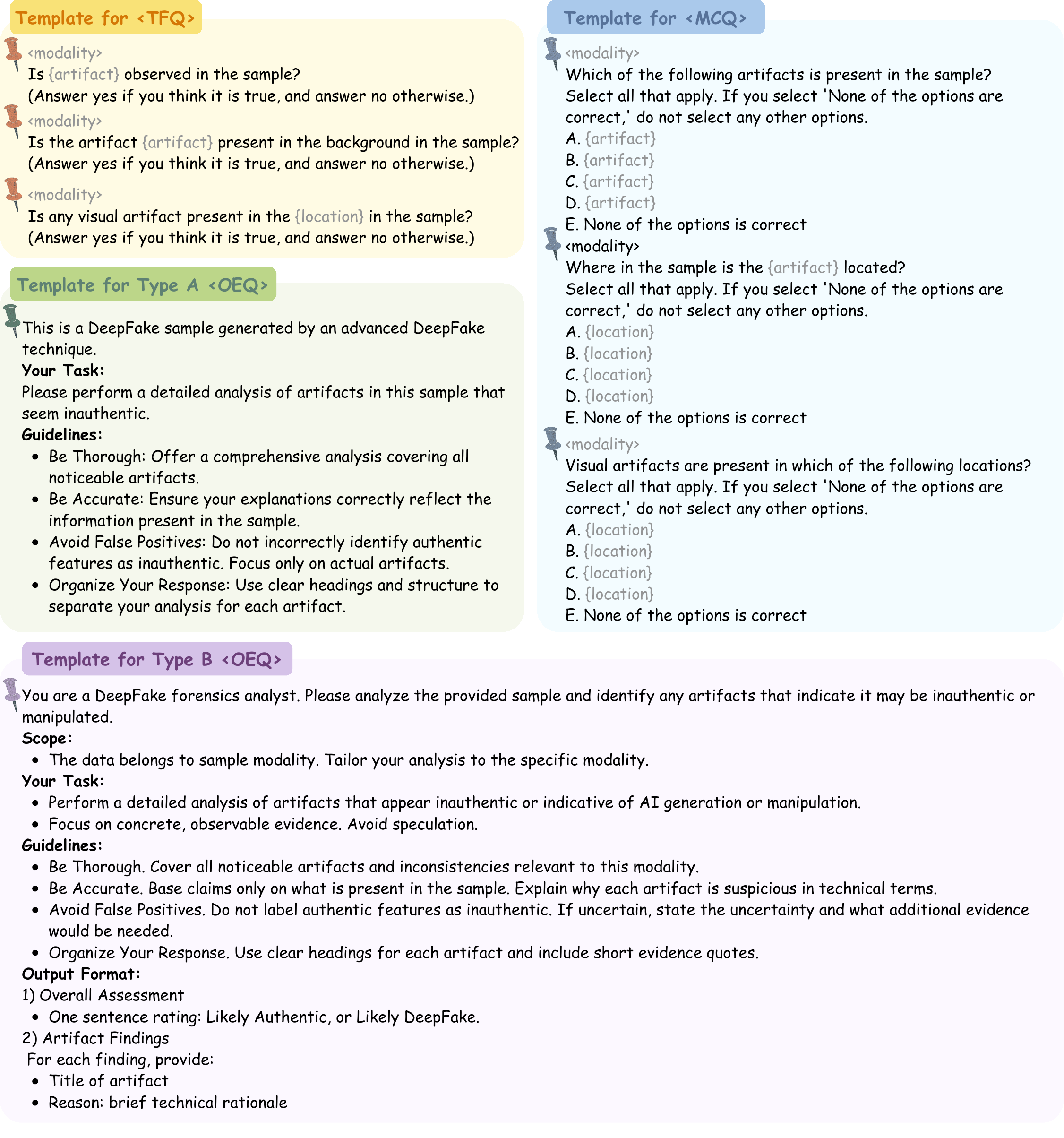}
  \caption{Prompt Template Used for Benchmark Construction for \texttt{<TFQ>}, \texttt{<MCQ>}, and \texttt{<OEQ>}}
  \label{fig:template_bench}
\end{figure}

\hky{\cref{fig:template_bench} outlines prompt templates designed for benchmark construction across three distinct question formats: \texttt{<TFQ>}, \texttt{<MCQ>}, and \texttt{<OEQ>}. The \texttt{<TFQ>} (True-False Question) section provides templates to verify the observation of specific artifacts, their presence in the background, or their existence in specific locations. The \texttt{<MCQ>} (Multiple-Choice Question) templates ask MLLMs to identify present artifacts or their locations from a list, including instructions to select all that apply or indicate if no options are correct. Finally, the \texttt{<OEQ>} (Open-Ended Question) templates, split into Type A and Type B, establish a persona for a DeepFake forensics analyst, detailing strict guidelines for performing thorough artifact analysis, avoiding false positives, and adhering to a specific output format.}

% \subsection{Templates for Similarity-based Evaluation}
% \label{sec:temp_eval}

\subsection{Templates for Artifacts Mapping}
\label{sec:temp_art_map}

\begin{figure}[t]
  \centering
  \includegraphics[width=\linewidth]{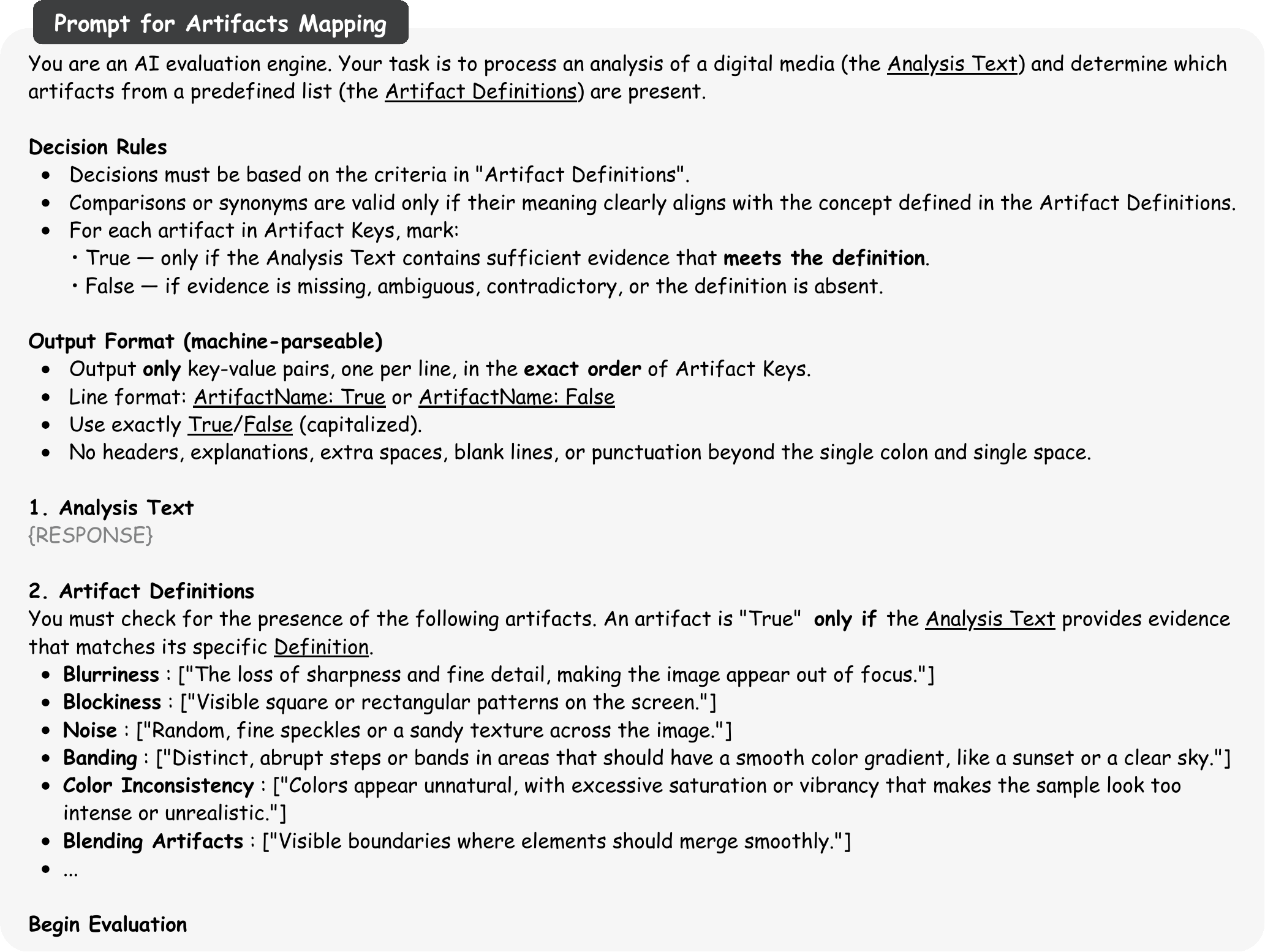}
  \caption{Prompt Template Used for Artifacts Mapping}
  \label{fig:template_mapping}
\end{figure}

\hky{\cref{fig:template_mapping} serves as a structured guide for identifying particular visual flaws in media analysis texts. It offers precise definitions of various artifacts as a reference point, compelling LLMs to assess their occurrence based on these exact standards. The template requires LLMs to deliver straightforward binary judgments of ``True'' or ``False,'' formatted in a machine-readable style using only key-value pairs.}

\section{Audio Modality Analysis}
\label{sec:audio}

\noindent{\textbf{Evaluation of Perception.}} \cref{tab:audio-perception} presents the audio perception performance of five open-weight Audio-MLLMs and one proprietary multimodal model. Two distinct trends emerge from the results. 

Firstly, semantic perception is substantially more challenging than quality perception. On \texttt{<TFQ>}, Gemini-2.5-Pro attains the highest semantic accuracy, yet most audio-specialized models perform near random chance in this regime. By contrast, these models often exhibit strong performance on quality-related artifacts. This divergence suggests that current systems still lean heavily on low-level signal cues rather than forming robust representations of prosody or speaker plausibility. A salient example is the semantic artifact of unnatural prosody: the waveform may appear clean, but subtle irregularities in rhythm, intonation, or stress make the speech sound implausible to human listeners. Such artifacts are notoriously hard for existing models to detect reliably, underscoring the intrinsic difficulty of semantic perception in audio.

Secondly, we hypothesize that this difficulty is partly driven by an architectural bias. Most MLLMs rely on audio encoders optimized for transcription or high-level semantic understanding, rather than for preserving speaker-identity fidelity or prosodic consistency. As a result, precisely those cues that are critical for judging who is speaking and whether their timing and intonation patterns are human-plausible are under-emphasized in the learned representations, limiting effective DeepFake perception in the audio modality.

\noindent{\textbf{Interpretable Detection, Perception and Hallucination.}}
We analyze interpretable audio deepfake detection using Type-A and Type-B \texttt{<OEQ>} questions, with full results summarized in ~\cref{tab:audio-open-ended}. For Type-A \texttt{<OEQ>}, only Qwen3-Omni-30B-A3B and Gemini-2.5-Pro produce meaningful artifact-level explanations. Qwen3-Omni achieves the highest \textit{Cover} and $F_{0.5}$ scores, albeit with a moderate level of hallucination, whereas Gemini-2.5-Pro attains slightly lower \textit{Cover} and $F_{0.5}$ scores but produces more consistently grounded descriptions. By contrast, audio-focused models such as Qwen2-Audio-7B, SALMONN-7B, and audio-flamingo-3 yield very low \textit{Cover} and near-saturated hallucination rates, resulting in almost zero $F_{0.5}$ scores. These findings indicate that current audio MLLMs still struggle to provide faithful artifact-level explanations and often hallucinate nonexistent distortions.

Type-B \texttt{<OEQ>} highlights a significant disparity between detection accuracy and explanation quality. SALMONN-7B achieves the highest detection accuracy but offers almost no interpretability, often providing the correct label while generating unreliable explanations. In contrast, Gemini 2.5-Pro demonstrates the opposite trend: its detection accuracy is nearly at chance levels, yet it provides the best interpretability, characterized by the highest \textit{Cover}, reduced hallucination, and the strongest $F_{0.5}$ score. Qwen3-Omni-30B-A3B and Phi-4 fall somewhere in between, exhibiting moderate accuracy and $F_{0.5}$ scores, but still suffering from considerable hallucination. Meanwhile, audio-flamingo-3 performs poorly in both detection and interpretability.

Overall, the audio results reinforce the main tri-perspective conclusion that current models rarely achieve both strong detection and low hallucination in this modality. Audio-centric MLLMs often depend on unclear heuristics and provide explanations that are highly prone to hallucination, whereas stronger multimodal models offer more grounded reasoning but show only slight improvements over random guessing. These findings highlight the need for better speech-specific perception modules and enhanced modeling of prosody and identity cues to achieve more reliable audio DeepFake detection.

\begin{table}[!t]
\centering
\caption{Evaluation of Audio Deepfake Perception}
\label{tab:audio-perception}
\renewcommand{\arraystretch}{1.}
\fontsize{6pt}{8pt}\selectfont
{\setlength{\tabcolsep}{0.6mm}
\resizebox{\columnwidth}{!}{
\begin{tabular}{l|cccc|cc}
\toprule
\multirow{2}{*}{\textbf{MLLM}}
& \multicolumn{4}{c|}{\textbf{\texttt{<TFQ>}}}
& \multicolumn{2}{c}{\textbf{\texttt{<MCQ>}}} \\
\cmidrule(lr){2-5}\cmidrule(lr){6-7}
& \emph{Semantic} & \emph{Quality} & \emph{Avg.} & \textbf{Rank}
& \emph{General} & \textbf{Rank} \\
\midrule
Random Guess             & 50.00\% & 50.00\% & 50.00\% & -- & \mc{0.00} & -- \\
\midrule
Qwen2-Audio-7B           & 44.50\% & 67.88\% & 56.19\% & 2 & \mc{0.01}  & 3 \\
Qwen3-Omni-30B-A3B       & 32.76\% & 67.37\% & 50.07\% & 3 & \mc{-0.15} & 5 \\
%SALMONN-7B               & 0.00\%  & 0.00\%  & 0.00\%  & 6 & \mc{0.00}  & 4 \\
Phi-4                    & 5.50\%  & 68.45\% & 36.98\% & 5 & \mc{-0.06} & 4 \\
Audio-Flamingo-3         & 6.91\%  & 67.88\% & 37.40\% & 4 & \mc{0.10}  & 1 \\
\midrule
Gemini-2.5-pro           & 63.65\% & 50.13\% & 56.89\% & 1 & \mc{0.04}  & 2 \\
\midrule
\textbf{Average} & 30.66\% & 64.34\% & 47.51\% & -- & \mc{-0.01} & -- \\
\bottomrule
\end{tabular}}}

\begin{tablenotes}[para,flushleft]
\scriptsize
\parbox{\columnwidth}{
%\textbf{Notes.} This table reports audio-only DeepFake perception performance. ``--'' indicates that the rank is not applicable.
}
\end{tablenotes}
\end{table}

\begin{table}[!t]
\centering
\caption{Evaluation of Interpretable Audio Deepfake Detection, Perception and Hallucination Robustness}
\label{tab:audio-open-ended}
\renewcommand{\arraystretch}{1.}
\fontsize{6pt}{8pt}\selectfont
{\setlength{\tabcolsep}{0.6mm}
\resizebox{\columnwidth}{!}{
\begin{tabular}{l|cccc|ccccc}
\toprule
\multirow{3}{*}{\textbf{MLLM}} 
& \multicolumn{4}{c|}{\textbf{\texttt{Type A <OEQ>}}} 
& \multicolumn{5}{c}{\textbf{\texttt{Type B <OEQ>}}} \\
\cmidrule(lr){2-5}\cmidrule(lr){6-10}
& \multicolumn{4}{c|}{\textbf{Audio}} 
& \multicolumn{5}{c}{\textbf{Audio}} \\
\cmidrule(lr){2-5}\cmidrule(lr){6-10}
& \emph{Cover} $\uparrow$ & \emph{CHAIR} $\downarrow$ & \emph{Hal} $\downarrow$ & $\mathbf{F}^{0.5}$ $\uparrow$
& \emph{ACC} & \emph{Cover} $\uparrow$ & \emph{CHAIR} $\downarrow$ & \emph{Hal} $\downarrow$ & $\mathbf{F}^{0.5}$ $\uparrow$ \\
\midrule
Qwen2-Audio-7B & 0.0446 & 0.9342 & 0.9421 & 0.0580 & 0.3799 & 0.2356 & 0.6388 & 0.6756 & 0.3166 \\
Qwen3-Omni-30B-A3B & \textbf{0.5278} & \textbf{0.2011} & \textbf{0.2867} & \textbf{0.7031} & 0.4082 & \underline{0.3690} & 0.5279 & 0.6756 & 0.4312 \\
SALMONN-7B & 0.0012 & 0.9973 & 0.9973 & 0.0021 & \textbf{0.5722} & 0.0472 & 0.9225 & 0.9225 & 0.0673 \\
Phi-4 & 0.1983 & 0.7375 & 0.7736 & 0.2360 & 0.3949 & 0.3185 & \underline{0.4967} & \underline{0.5187} & \underline{0.4398} \\
Audio-Flamingo-3 & 0.0811 & 0.8708 & 0.8708 & 0.1129 & 0.3732 & 0.0206 & 0.9635 & 0.9635 & 0.0311 \\
\midrule
Gemini 2.5-pro & \underline{0.3065} & \underline{0.5079} & \underline{0.5339} & \underline{0.4279} & \underline{0.4859} & \textbf{0.5470} & \textbf{0.2106} & \textbf{0.2736} & \textbf{0.7022} \\
\bottomrule
\end{tabular}}}

\end{table}

\section{Extended Evaluation}
\label{sec:exp_more}

\subsection{Evaluation Setup}
%\noindent\textbf{Evaluation models and modalities.} \hky{We benchmark $22$ MLLMs ($19$ open-source, $3$ proprietary) across image, video, and audio modalities. For visual tasks, open-source models include InternVL2\_5/3\_5~\cite{chen2024expanding,wang2025internvl3}, Qwen3-Omni/VL~\cite{xu2025qwen3,bai2025qwen3}, LLaVA-OV~\cite{li2024llava}, MiniCPM-V~\cite{yao2024minicpm}, MiMo-VL~\cite{yue2025mimo}, Idefics2~\cite{laurenccon2024matters}, Mantis~\cite{jiang2024mantis}, Phi-4~\cite{abdin2024phi}, and the forensic-focused FakeShield~\cite{xu2025fakeshield}. These are compared against proprietary baselines: GPT-5~\cite{gpt5}, Gemini 2.5-Pro~\cite{comanici2025gemini}, and Claude Sonnet 4.5~\cite{Claude}. Audio performance is evaluated using Qwen2-Audio~\cite{chu2024qwen2}, Qwen3-Omni, Phi, Audio-Flamingo-3~\cite{ghosh2025audio}, and SALMONN-7B~\cite{tang2024salmonn}, with Gemini 2.5-Pro serving as the proprietary reference.}

\noindent\textbf{Evaluation models and modalities.} 
For visual modalities, we consider open-source MLLMs including InternVL2\_5/3\_5~\cite{chen2024expanding,wang2025internvl3}, Qwen3-Omni/VL~\cite{xu2025qwen3,bai2025qwen3}, LLaVA-OV~\cite{li2024llava}, MiniCPM-V~\cite{yao2024minicpm}, MiMo-VL~\cite{yue2025mimo}, Idefics2~\cite{laurenccon2024matters}, Mantis~\cite{jiang2024mantis}, Phi-4~\cite{abdin2024phi}, and the forensic-focused FakeShield~\cite{xu2025fakeshield} and FakeVLM~\cite{wen2025spot}. 
These are compared against proprietary baselines: GPT-5~\cite{gpt5}, Gemini 2.5-Pro~\cite{comanici2025gemini}, and Claude Sonnet 4.5~\cite{Claude}. Audio performance is evaluated using Qwen2-Audio~\cite{chu2024qwen2}, Qwen3-Omni, Phi, Audio-Flamingo-3~\cite{ghosh2025audio}, and SALMONN-7B~\cite{tang2024salmonn}, with Gemini 2.5-Pro serving as the proprietary reference.

\noindent\textbf{Experimental protocol.}
All experiments are conducted in a zero-shot setting, where each sample is processed independently without task-specific fine-tuning. For each query, we provide the model with the question prompt together with the corresponding image, video, or audio input. For video tasks, we either use a 16-frame clip (when frame sampling is configurable) or the model's default frame sampling policy. 
Unless otherwise noted, the same protocol is applied consistently across all models and modalities.

\subsection{More Quantitative Results}
\label{sec:exp_quant_more}
\noindent{\textbf{Interplay between perception, hallucination, and detection.}}
To understand how the three evaluation dimensions of \benchname\ relate to one another, we analyze the correlations between perception, hallucination and detection performances across all $22$ evaluated models. For each model \(m\), we compute three macro-averaged scores over all available samples: (i) perception \(P_m\), defined as Type-A \textit{Cover}; (ii) hallucination severity \(H_m\), defined as Type-A \textit{CHAIR}; and (iii) detection \(D_m\), defined as Type-B \texttt{<OEQ>} detection accuracy. 

\begin{figure}[t]
  \centering
  \includegraphics[width=\linewidth]{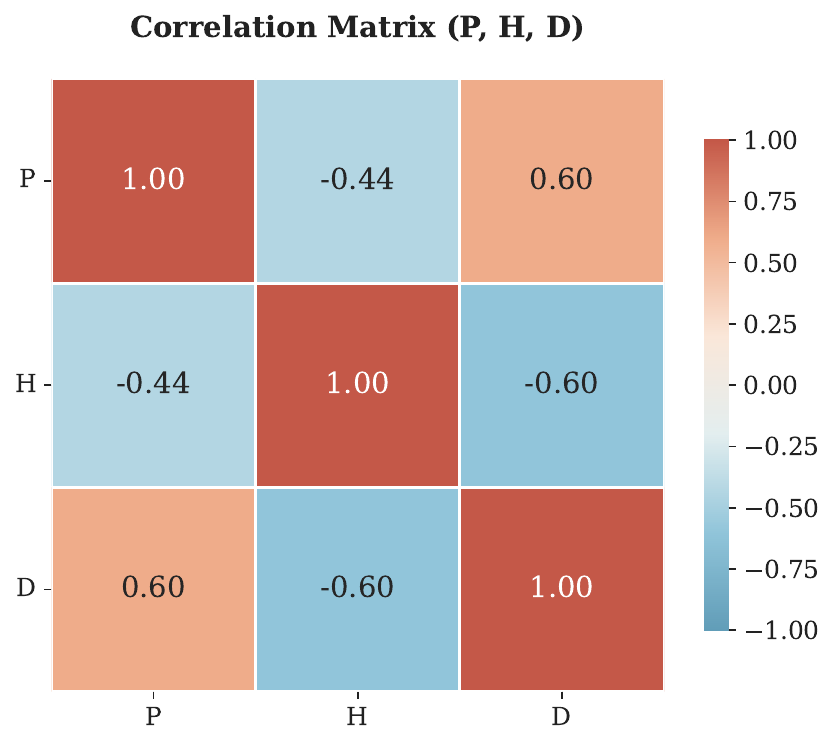}
  \caption{Model-level correlation matrix for perception (P), hallucination severity (H), and detection (D). Perception is positively correlated with detection accuracy, while hallucination is negatively correlated with both, supporting the three-dimensional P–H–D view of MLLM-based DeepFake detection.}
  \label{fig:correlation}
\end{figure}

The resulting correlation matrix in~\cref{fig:correlation} reveals a tightly coupled but non-degenerate triad. Perception and detection are moderately positively correlated \((r(P,D)\approx 0.60)\): models that cover more ground-truth artifacts in Type-A explanations tend to achieve higher Type-B detection accuracy. Hallucination severity is also strongly coupled to detection \((r(H,D)\approx -0.60)\), with more hallucinated artifacts associated with lower accuracy. Although perception and hallucination are negatively correlated \((r(P,H)\approx -0.44)\), the magnitude of this correlation is relatively moderate. It indicates that while models that recognize more genuine artifacts tend to hallucinate less, the two aspects remain far from interchangeable.
The overall correlation matrix shows that perception and detection are moderately aligned, while hallucination undermines detection and is moderately anti-correlated with perception. 

\begin{figure}[t]
  \centering
  \includegraphics[width=\linewidth]{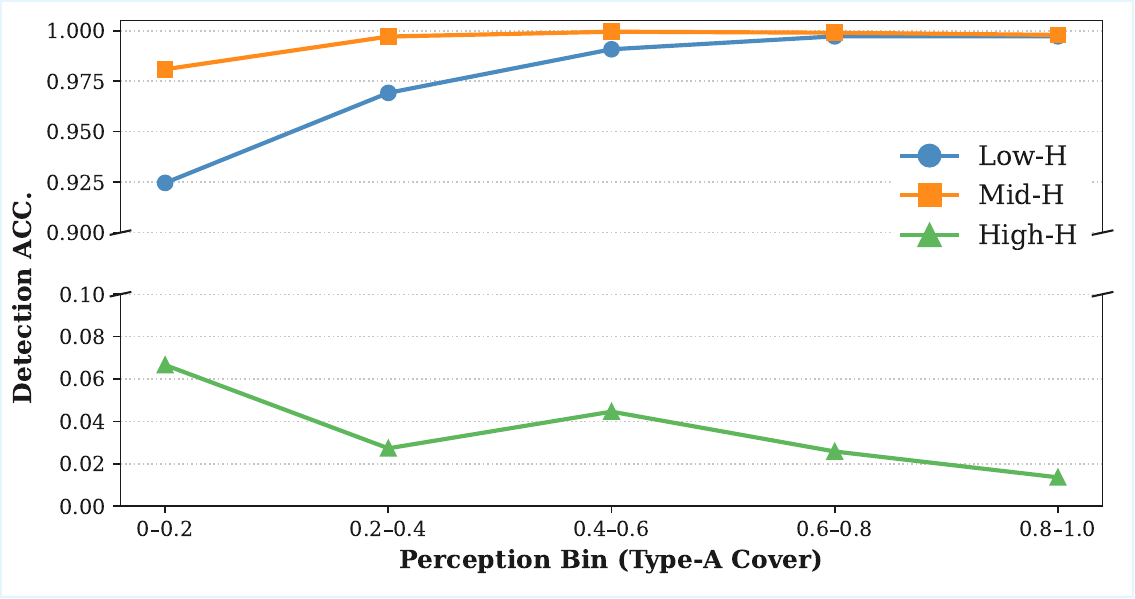}
    \caption{Stratified perception–detection curves on TriDF: fake-detection accuracy vs. binned Type-A \textit{Cover} under three Type-A \textit{CHAIR} regimes, showing that strong hallucination keeps detection near chance even with high perceptual coverage.}
  \label{fig:stratified_plot}
\end{figure}

However, when we further stratify models by hallucination severity, a more revealing pattern emerges. We define hallucination regimes using the empirical sample distribution: all samples with \(H=1\) form a high-hallucination regime (High-H), while samples with \(H<1\) are split at the 33rd and 67th percentiles into Low-H and Mid-H, and analyze the fake-only subset of TriDF. Independently, we discretize perception into five equal-width bins based on Type-A \textit{Cover} (\(0\text{--}0.2, 0.2\text{--}0.4, \dots, 0.8\text{--}1.0\)). For each hallucination regime and perception bin, we then compute the average fake detection accuracy and plot the resulting curves in ~\cref{fig:stratified_plot}.

The stratified curves reveal a clear three-way interaction. In the Low-H and Mid-H regimes, fake-detection accuracy is high at low \textit{Cover} and rapidly saturates near perfect accuracy as \textit{Cover} increases, indicating that once explanations are largely grounded, additional perceptual coverage yields gains on detection accuracy. In contrast, in the High-H regime, DeepFake detection accuracy remains close to chance across all perception bins and is effectively insensitive to \textit{Cover}. Even when models capture numerous artifacts (high \(P\)), severe hallucination in Type-A explanations is associated with systematic failures to flag fakes in Type-B decisions. 

Both analyses shown in~\cref{fig:correlation} and~\cref{fig:stratified_plot} demonstrate that hallucination can disrupt the natural link between evidence recognition in perception and detection decision-making. The findings reinforce that perception, detection, and hallucination capture fundamentally distinct aspects of model behavior, and that reliable DeepFake detection requires balanced progress across all three dimensions. Improving only perception or only classification is insufficient. Addressing these intertwined but independent factors is crucial for building trustworthy and human-aligned detection systems capable of withstanding increasingly sophisticated forgeries.

\noindent \textbf{Benefit-Cost Analysis of Localization Hints. }
\hky{As discussed in RQ2 in the main paper, we quantify the efficacy of localization hints and define \textit{Benefit} and \textit{Cost} as the percentages of questions where the hint respectively corrects an initial error or induces a new one. Their difference, \textit{Net Benefit}, serves as the primary indicator of genuine performance gain from spatial guidance.}
\hky{The results are summarized in~\cref{tab:rq2_benefit_cost}. Localization hints generally yield a positive \textit{Net Benefit}, though gains vary by architecture. InternVL2\_5-8B and Claude Sonnet 4.5 achieve peak efficiency ($2.53$\% and $2.47$\% \textit{Net Benefit}), demonstrating an effective ability to leverage spatial cues. Conversely, Gemini 2.5-Pro and Qwen3-VL-30B-Instruct exhibit negative \textit{Net Benefit} ($-0.30$\% and $-0.32$\%), suggesting that for certain high-capacity architectures, external hints may introduce disruptive noise. This non-universal efficacy underscores a persistent architectural gap in reconciling external spatial grounding with internal visual representations.}

\begin{table}[!ht]
\centering
\caption{RQ2. Benefit and Cost of localization hints.}
\label{tab:rq2_benefit_cost}
\renewcommand{\arraystretch}{1.}
\fontsize{6pt}{8pt}\selectfont
{\setlength{\tabcolsep}{0.6mm}
\resizebox{\columnwidth}{!}
{\begin{tabular}{lcccc}
\toprule
\textbf{MLLM} & \textit{Benefit} (\%) & \textit{Cost} (\%) & \textit{Net Benefit} (\%) & \textbf{Rank}\\
\midrule
InternVL2\_5-8B~\cite{chen2024expanding}       & 3.21  & 0.68  & \textbf{\mc{2.53}}  & 1  \\
InternVL2\_5-26B~\cite{chen2024expanding}      & 4.28  & 1.90  & \mc{2.38}  & 4  \\
InternVL2\_5-38B~\cite{chen2024expanding}      & 4.22  & 1.78  & \mc{2.44}  & 3  \\
InternVL3\_5-8B~\cite{wang2025internvl3}       & \underline{10.87} & \underline{10.10} & \mc{0.78}  & 7  \\
InternVL3\_5-38B~\cite{wang2025internvl3}      & 4.93  & 3.30  & \mc{1.63}  & 5  \\
Qwen3-Omni-30B-A3B-Instruct~\cite{xu2025qwen3} & 6.92  & 6.55  & \mc{0.37}  & 9  \\
Qwen3-VL-8B-Instruct~\cite{bai2025qwen3}       & 8.03  & 7.30  & \mc{0.73}  & 8  \\
Qwen3-VL-30B-Instruct~\cite{bai2025qwen3}      & 7.01  & 7.33  & \mc{-0.32} & 11 \\
GPT-5~\cite{gpt5}                              & 6.57  & 5.65  & \mc{0.92}  & 6  \\
Gemini 2.5-Pro~\cite{comanici2025gemini}       & \textbf{11.67} & \textbf{11.97} & \mc{-0.30} & 10 \\
Claude Sonnet 4.5~\cite{Claude}                & 3.17  & 0.70  & \textbf{\mc{2.47}}  & 2  \\
\bottomrule
\end{tabular}}}
\end{table}

\textbf{}

\subsection{More Qualitative Results}
\label{sec:exp_qual_more}

\hky{Based on the provided documents, the case studies utilize three distinct evaluation formats, \texttt{<TFQ>}, \texttt{<MCQ>}, and \texttt{<OEQ>}, to assess model performance in detecting synthesis and manipulation artifacts.}

\begin{figure*}[!ht]
  \centering
  \includegraphics[width=\linewidth]{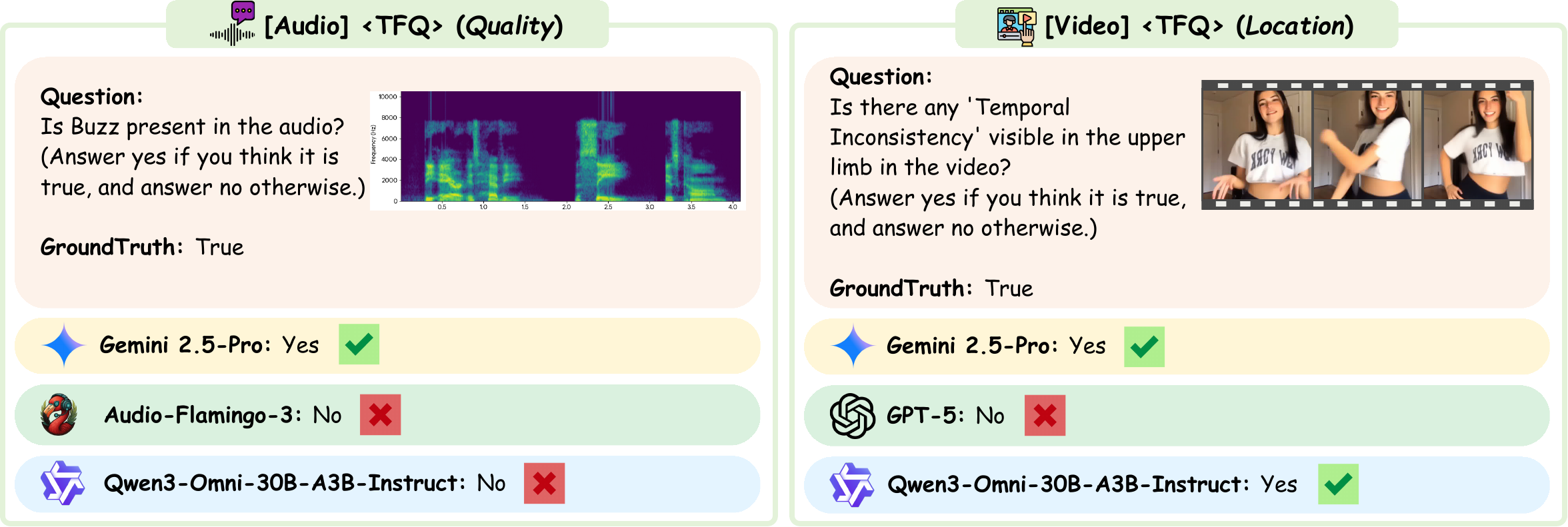}
  \caption{Examples of \texttt{<TFQ>}}
  \label{fig:vis_tfq}
\end{figure*}

\noindent \texttt{<TFQ>} focuses on binary verification, prompting models to simply confirm or deny the presence of specific defects, such as detecting ``Buzz'' in an audio clip or identifying ``Temporal Inconsistency'' in a video subject's upper limb. As shown in~\cref{fig:vis_tfq}, Gemini 2.5-Pro outperforms both powerful general-purpose model (\eg, Qwen3-Omni-30B-A3B-Instruct) and specialized model, Audio-Flamingo-3. Conversely, GPT-5 struggles in this example because it cannot handle raw video inputs without preprocessing, which hinders its ability to understand temporal relationships.

\begin{figure*}[!ht]
  \centering
  \includegraphics[width=\linewidth]{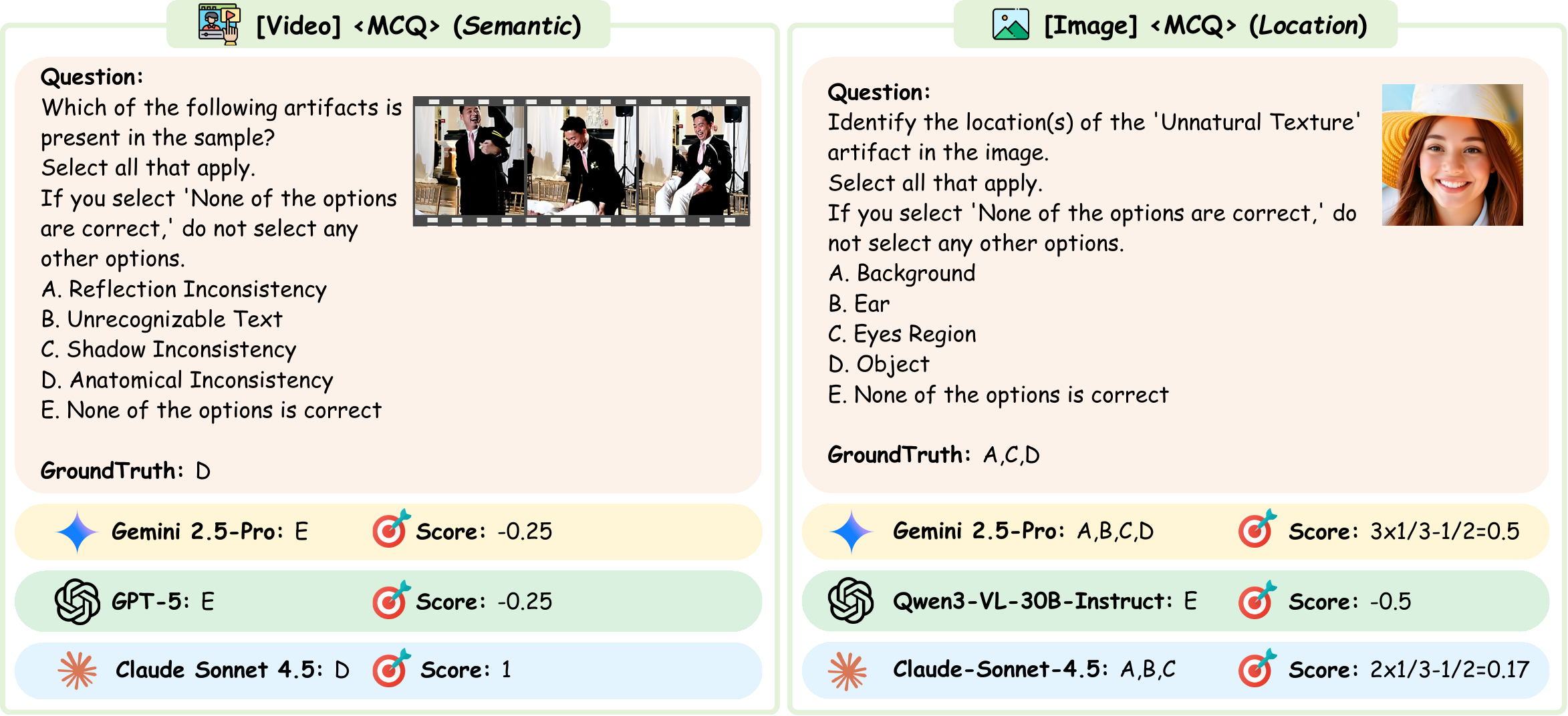}
  \caption{Examples of \texttt{<MCQ>}}
  \label{fig:vis_mcq}
\end{figure*}

\noindent \texttt{<MCQ>} tests the ability to categorize or locate specific errors, asking models to identify semantic issues like ``Anatomical Inconsistency'' or select specific regions where artifacts appear, such as the ``Ear'' or ``Background''. Within the two examples in~\cref{fig:vis_mcq}, the evaluation metric is strict: models must answer all options correctly to receive the maximum score of $1$. Any incorrect selection results in a penalty, preventing a full score.

\begin{figure*}[!ht]
  \centering
  \includegraphics[width=\linewidth]{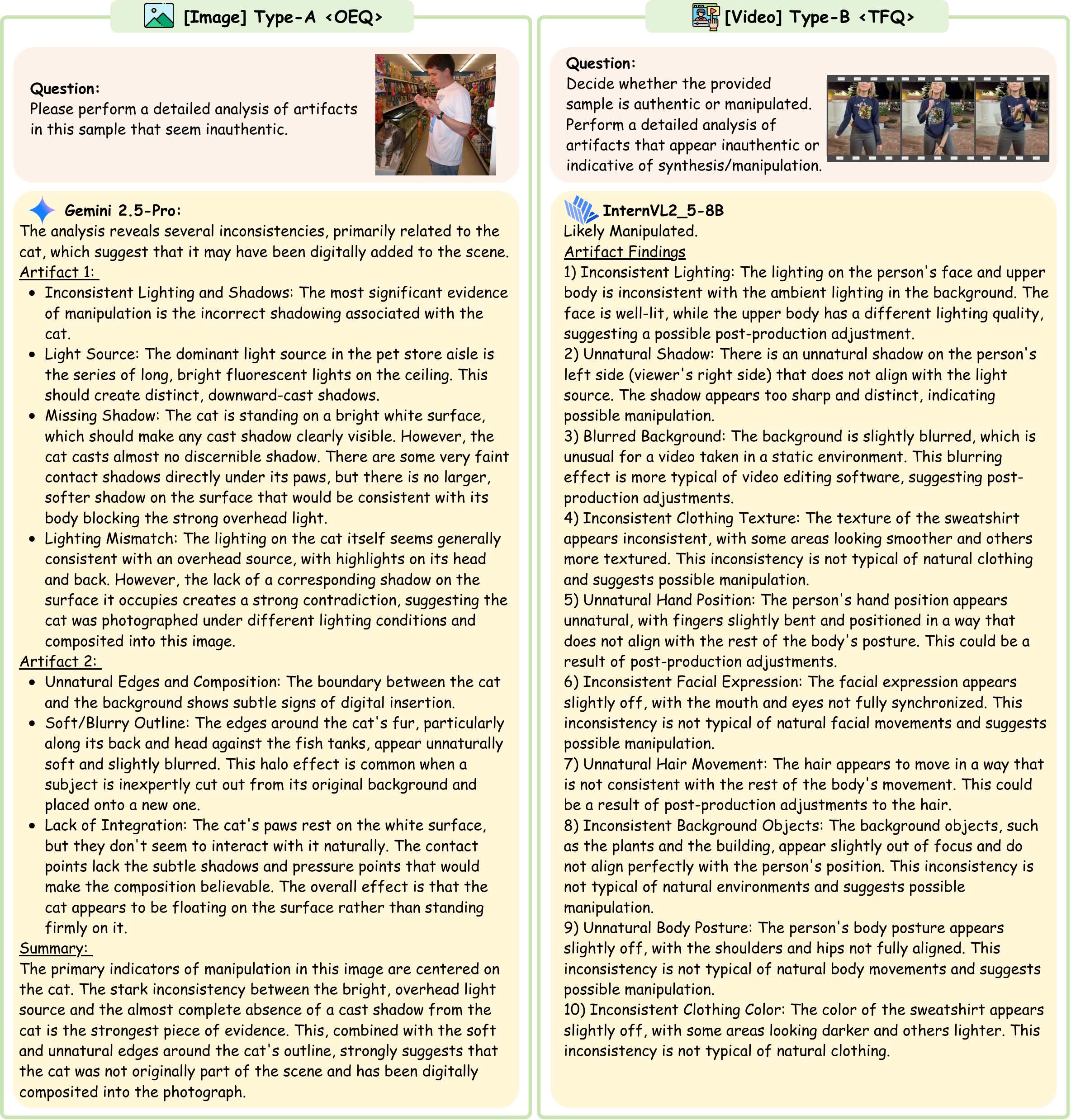}
  \caption{Examples of \texttt{<OEQ>}}
  \label{fig:vis_oeq}
\end{figure*}

\noindent Finally, \texttt{<OEQ>} requires a more granular, descriptive analysis, asking models to justify a ``Likely Manipulated'' verdict by detailing observable flaws like ``Inconsistent Lighting'', ``Unnatural Shadow'', or a ``Blurred Background''.~\cref{fig:vis_oeq} highlights the variance in model perspective: Gemini 2.5-Pro provides a focused, context-aware analysis of lighting physics on a specific object (a cat), whereas InternVL2\_5-8B generates a generic list of DeepFake flaws typically associated with human subjects.

\section{Future Direction of DeepFake Detection}
\label{sec:future_work}
\lynn{\benchname\ fills an important gap in existing evaluation resources by enabling systematic analysis of all three components. Looking forward, \benchname\ provides several avenues for advancing future DeepFake detection techniques. First, the fine-grained artifact taxonomy offers a structured supervisory signal that can guide new models to focus on meaningful manipulation cues rather than dataset-specific shortcuts. Second, the multimodal and diverse generator design creates a challenging testbed that encourages the development of detectors with stronger generalization across synthesis pipelines. Third, the hallucination evaluation reveals failure modes in explanation generation and provides a foundation for designing models that produce grounded, reliable reasoning. Finally, as new generative techniques and modalities emerge, \benchname\ can be extended to support evolving research needs, serving as a long-term platform for building trustworthy and deployable DeepFake detection systems.}

\section{Release Plan and Ethics Statement}
\hky{All datasets utilized in this benchmark are sourced from publicly available repositories. DeepFake generation was conducted strictly for academic and research purposes to advance the fields of media forensics and authenticity detection. Our research team explicitly opposes the malicious application of this technology and condemns any use of this benchmark or the associated data for deceptive, harmful, or misinformation-related purposes.}